\pdfoutput=1
\documentclass[11pt]{article}

\usepackage[final]{acl}

\usepackage{times}
\usepackage{latexsym}
\usepackage{todonotes}
\usepackage{multirow}
\usepackage{graphicx}
\usepackage{amsmath}
\usepackage{amsfonts}
\usepackage{amssymb}
\usepackage{caption}
\usepackage[T1]{fontenc}
\usepackage{pifont}
\usepackage{subcaption}
\usepackage{booktabs}
\usepackage{colortbl}
\usepackage{amsmath} 

\usepackage{enumitem}
\usepackage{amssymb}
\usepackage{pifont}
\usepackage{graphicx}

\newcommand{\cmark}{\color[HTML]{32CB00} \ding{51}} % Check mark
\newcommand{\xmark}{\color[HTML]{FE0000} \ding{55}} % Cross mark

\usepackage[utf8]{inputenc}

\usepackage{microtype}

\usepackage{inconsolata}
\usepackage{hyperref}
\title{\texttt{DocEdit-v2}: Document Structure Editing Via Multimodal LLM Grounding}

\author{Manan Suri $^{\text{\text{\ding{168}}}}$, Puneet Mathur$^{\text{\ding{171}}}$, Franck Dernoncourt$^{\text{\ding{171}}}$, \\ \textbf{Rajiv Jain$^{\text{\ding{171}}}$, Vlad I. Morariu$^{\text{\ding{171}}}$, Ramit Sawhney$^\text{\ding{169}}$, Preslav Nakov$^\text{\ding{169}}$, Dinesh Manocha$^{\text{\ding{168}}}$}\\
$^{\text{\ding{171}}}$Adobe Research, $^\text{\ding{169}}$ MBZUAI, Abu Dhabi, $^{\text{\ding{168}}}$University of Maryland College Park \\
{\tt manans@umd.edu}, 
{\tt puneetm@adobe.com}}

\begin{document}
\maketitle

\begin{abstract}
Document structure editing involves manipulating localized textual, visual, and layout components in document images based on the user's requests. Past works have shown that multimodal grounding of user requests in the document image and identifying the accurate structural components and their associated attributes remain key challenges for this task. To address these, we introduce the \texttt{DocEdit-v2}, a novel framework that performs end-to-end document editing by leveraging Large Multimodal Models (LMMs). It consists of three novel components -- (1) Doc2Command to simultaneously localize edit regions of interest (RoI) and disambiguate user edit requests into edit commands. (2) LLM-based Command Reformulation prompting to tailor edit commands originally intended for specialized software into edit instructions suitable for generalist LMMs. (3) Moreover, \texttt{DocEdit-v2} processes these outputs via Large Multimodal Models like GPT-4V and Gemini, to parse the document layout, execute edits on grounded Region of Interest (RoI), and generate the edited document image. Extensive experiments on the DocEdit dataset show that \texttt{DocEdit-v2} significantly outperforms strong baselines on edit command generation (2-33\%), RoI bounding box detection (12-31\%), and overall document editing (1-12\%) tasks.
\end{abstract}

\section{Introduction}

\begin{figure}[h!]
  \centering
  \includegraphics[width=0.8\columnwidth]{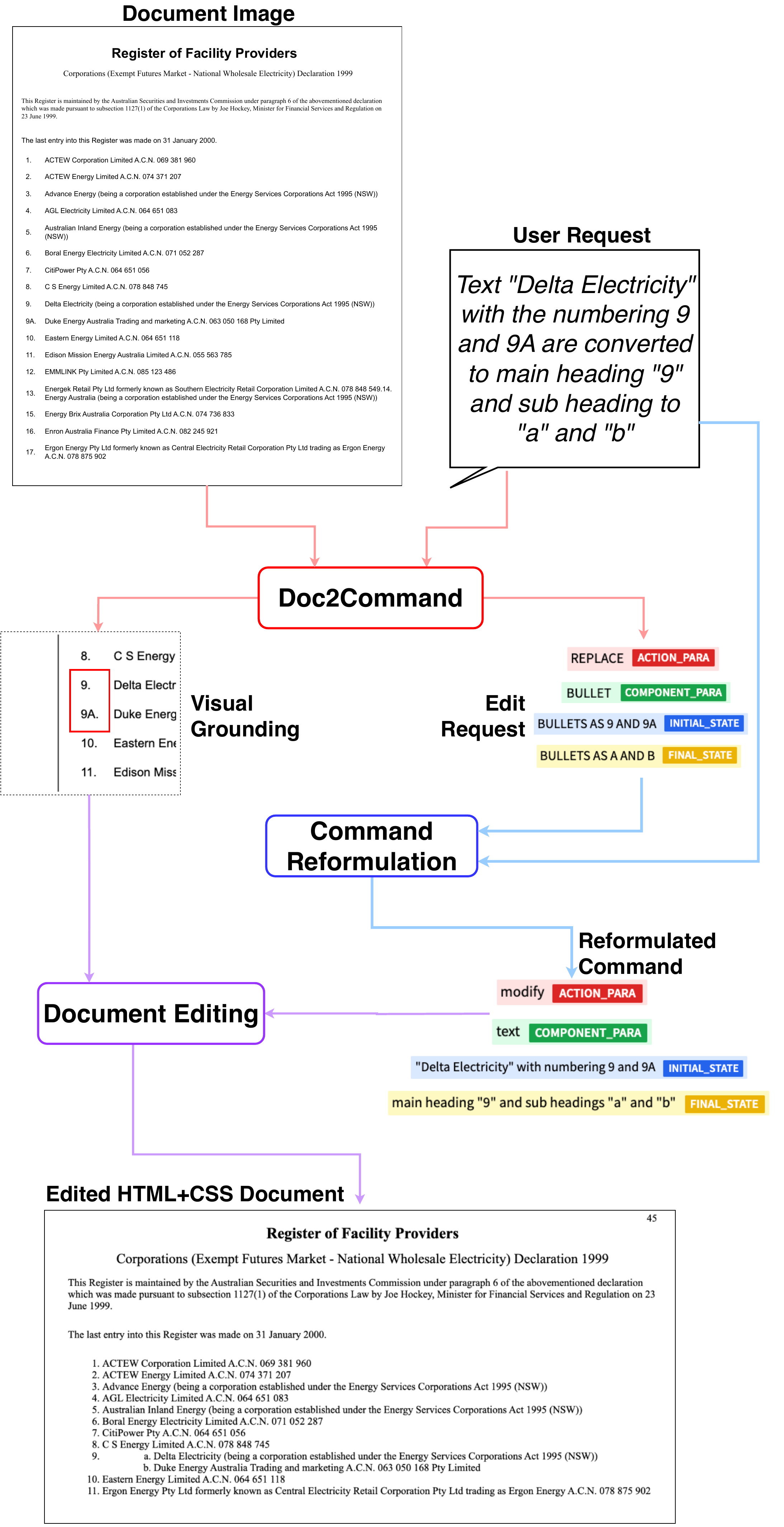}
  \caption{\texttt{DocEdit-v2} framework performs multimodal grounding and edit command generation via Doc2Command, utilizes LLM-based Command Reformulation prompting to refine the command into  LMM instruction format (<\colorbox{red}{Action}><\colorbox{green}{Component}>, 
  % <\colorbox{cyan}{Attribute}>, 
  <\colorbox{blue}{Initial State}>, <\colorbox{yellow}{Final State}>), and employs LMMs to edit the HTML structure using multimodal (edit instruction and grounded RoI) prompt.}
  \label{fig:docedit-intro}
\end{figure}
Digital documents are widely used for communication, information dissemination, and business productivity. Language-guided Document Editing entails modifying the textual, visual, and structural components of a document in response to a user's open-ended requests related to spatial alignment, component placement, regional grouping, replacement, resizing, splitting, merging, and applying special effects \cite{mathur2023docedit, Kudashkina2020DocumenteditingAA}. Document editing is inherently a generative task as it involves the creation of a new edited output from an existing document. 

\citet{mathur2023docedit} highlights three key challenges in the end-to-end document editing task -- (1) multimodal grounding of ambiguous user requests in the document image, (2) identifying the precise components and their corresponding attributes to be edited, and (3) generating faithful edits without distorting the semantic or spatial coherence of the original document. By interpreting the visual-semantic cues from user requests, multimodal grounding can bridge the gap between natural language instructions and the spatial intricacies of the document's content.  Sophisticated edit commands, like those found in the DocEdit dataset \cite{mathur2023docedit}, are usually ambiguous in nature and tailored for use in software-specific applications. Disambiguation of such edit commands can help to serve as refined editing instructions for generalist generation models. We hypothesize that directly editing the parsed HTML/XML document structure can overcome the limitations of pixel-level image generation. 

Prior works like DocEditor \citet{mathur2023docedit} performed edit commands generation for language-guided document editing but was limited to software-specific applications. Generative methods such as diffusion models have shown promise in the visual domain but pose challenges in recreating complex textual and visual elements while preserving the structural information of documents \cite{Yang2023DocDiffDE, He2023DiffusionbasedDL}. Unlike natural images, documents contain a combination of text, images, formatting, and layout intricacies \cite{mathur2023layerdoc} that necessitate a more nuanced approach to generative editing. Recently, Large Multimodal Models (LMMs) like GPT-4V \cite{OpenAI2023GPT4TR} and Gemini \cite{team2023gemini} have demonstrated remarkable capabilities in document understanding, object localization, dense captioning, and code synthesis. Prior work has also explored LLM program synthesis to compose vision-and-language queries into code subroutines \cite{Gao2022PALPL, Suris2023ViperGPTVI, Feng2023LayoutGPTCV, Huang2023Instruct2ActMM}. Our work aims to solve end-to-end editing of HTML representation of documents by leveraging the emergent capabilities of LMMs to infer the semantic context of edit requests, visually reference them to the region of interest in the document image, determine the spatial elements to be modified, and generate the final document.

\begin{figure*}[h!]
    \centering
    \includegraphics[width=\linewidth]{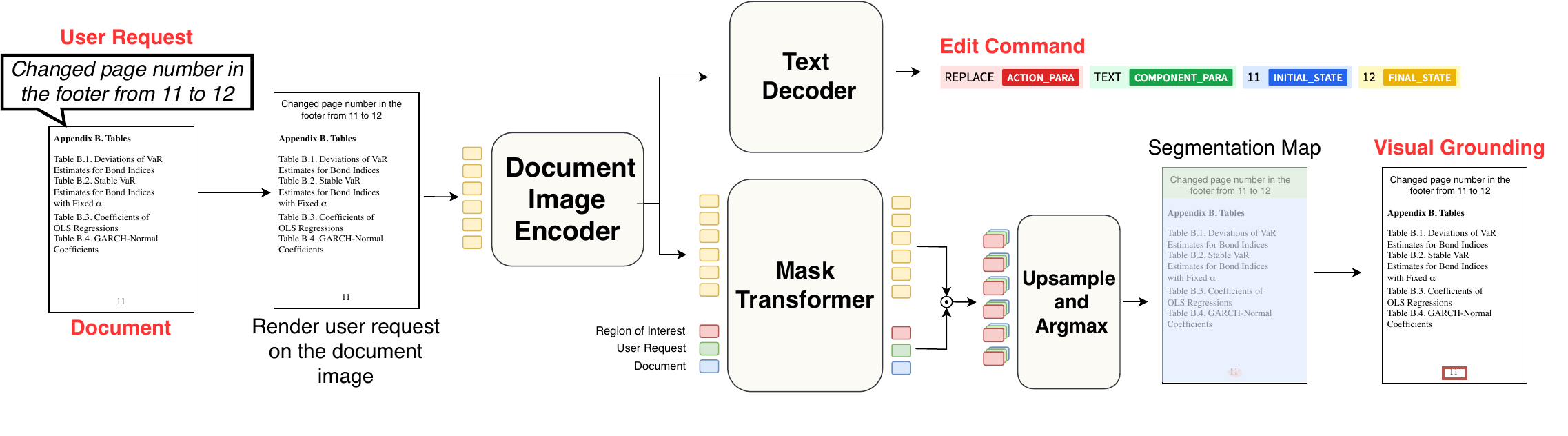}
    \caption{Doc2Command: Given a document image and a user request, the user request is rendered onto the document, and passed as a singular visual modality to an image encoder. The image encoder feeds into a text decoder and a mask transformer to generate the command text and segmentation maps, respectively.}
    \label{fig:Doc2Command}
\end{figure*}
\noindent\textbf{Main Results}: We present \texttt{DocEdit-v2} (Fig.\ref{fig:docedit-intro}) -- an LMM-based end-to-end document editing framework. Given a user request on a document, it utilizes a novel Doc2Command module to ground the edit location in the document image and generate edit commands. Doc2Command is a Transformer-based image encoder-text decoder-mask transformer model that is jointly trained to perform masked semantic segmentation and ground edit regions of interest (RoI) for disambiguating user edit requests into modularized commands. Doc2Command starts with visually integrating the edit request with the document image, processing them as a unified visual modality through a vision encoder-text decoder backbone to generate the command text. It redefines bounding box detection as a segmentation task by incorporating a mask-attention transformer over the image encoder. Further, we propose Command Reformulation prompting to customize the edit commands into an LMM-specific editing instruction by leveraging the zero-shot in-context learning ability of LLMs. Lastly, \texttt{DocEdit-v2} leverages LMMs such as GPT-4V and Gemini to edit the HTML structure of the document using a multimodal prompt formed by combining the edit instruction and grounded RoI. We design two new metrics - CSS IoU, and DOM Tree Edit Distance to evaluate the final edited documents for presentation quality and structural similarity with the ground truth. Experiments on the DocEdit dataset reveal that \textbackslash{}texttt\{DocEdit-v2\} significantly outperforms strong baselines in edit command generation (by 2-33\%), RoI bounding box detection (by 12-31\%), and overall document editing tasks (by 1-12\%).
% Experimental results demonstrate that DocEdit-v2 significantly outperforms baseline models with a 2-33\% improvement in command text generation , 12-31\% gain in bounding box detection task. 
Our \textbf{main contributions} are:
\begin{itemize}
\item We propose \textbf{Command Reformulation} to resolve ambiguity by using Large Language Models (LLMs) to translate the user's linguistic intent into a specific visual editing prompt for LMMs.
\item We introduce \textbf{Doc2Command}, a novel model for grounding edit requests that employs a transformer-based image encoder and text decoder architecture. It generates precise commands for document editing and semantically anchors editing regions through masked semantic segmentation in a multitask framework.
\item We present \textbf{DocEdit-v2}, an LMM-based framework for document editing. It interprets user requests to perform localized editing tasks conversationally. DocEdit-v2 utilizes Command Reformulation to convert user intent into appropriate LMM prompts and incorporates multimodal grounding via our proposed Doc2Command module.
\item Additionally, we define two new metrics - CSS IoU and DOM Tree Edit Distance - to assess LMM-generated documents for presentation quality and structural fidelity compared to ground truth.
\end{itemize}

\section{Related Work}

Past works in the domain of language-guided image editing have predominantly centered on natural image datasets \cite{image_dataset1, image_dataset2}, overlooking the distinctive characteristics of documents, which typically exhibit text-rich content alongside a diverse array of structured elements arranged in various layouts. These datasets often lack representations of localized edits and indirect edit references, crucial facets for effective document editing. Notably, contemporary GAN-based \cite{gan1, gan2, gan3, gan4, gan5} and diffusion methods \cite{diffusion1, diffusion2, diffusion3, diffusion4, diffusion5} have gained traction for natural image manipulation tasks due to their capacity for end-to-end pixel-level image synthesis. However, their applicability to digital documents, characterized by rich textual content and complex layouts, remains limited. These techniques are ill-equipped to grasp the spatial and semantic intricacies inherent in embedded textual components within documents. Consequently, prior endeavors in language-guided document editing have primarily pivoted towards multimodal grounding of edit requests through textual and visual cues into actionable commands and visual localization \cite{mathur2023docedit}. Despite these efforts, the absence of efficient generative frameworks tailored for document image editing remains a significant challenge in this domain. 

\begin{figure*}[h!]
    \centering
    \begin{subfigure}{\textwidth}
        \centering
        \includegraphics[width=0.9\textwidth]{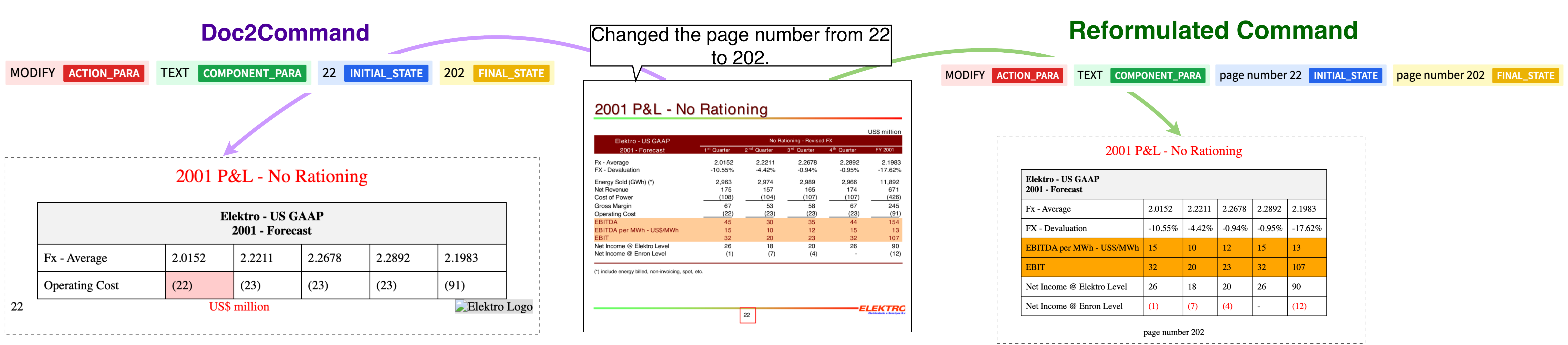}
        \caption{The reformulated command, by virtue of specificity, is able to achieve the desired edit.}
        \label{fig:eg_1}
    \end{subfigure}
    \hfill
    \begin{subfigure}{\textwidth}
        \centering
        \includegraphics[width=0.9\textwidth]{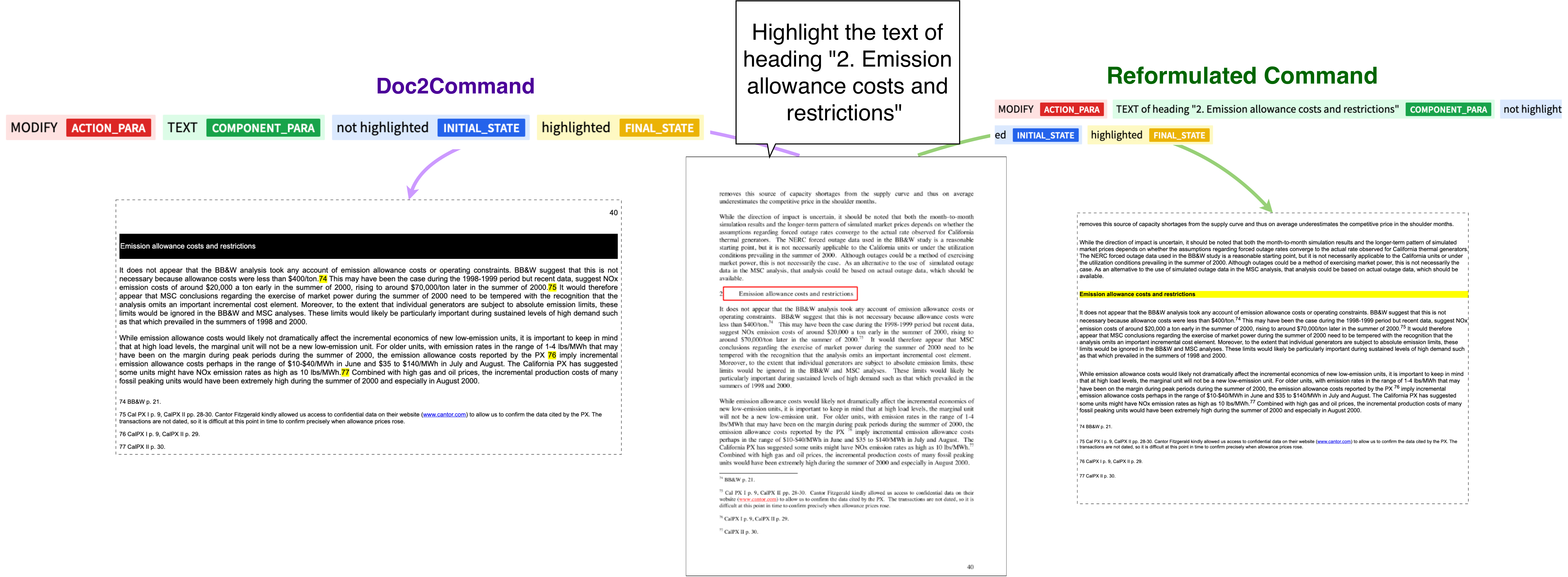}
        \caption{Commands reformulation performs better document grounding due to ambiguities in the generated command and distractors in the document image.}
        \label{fig:eg_2}
    \end{subfigure}
    \caption{Examples showing commands generated post-Doc2Command and Command Reformulation prompting.}
    \label{fig:both}
\end{figure*}

\section{\texttt{DocEdit-v2} Methodology}
\texttt{DocEdit-v2} (Fig.~\ref{fig:docedit-intro}) comprises of the following steps to ensure effective edit operation: (a) multimodal grounding and edit command generation via the Doc2Command, (b) Command Reformulation prompting to transform the edit command into LMM-specific prompt instruction, (c) prompting LMMs like GPT-4V and Gemini to facilitate nuanced and localized editing of the document's HTML representation.

%%%%%%%%

\subsection{Doc2Command}
Editing documents based on user requests requires converting open-vocabulary user requests into precise actions and grounding the region of interest in the document image. Edit command generation involves semantically mapping the ambiguous natural language user requests to specific editing actions, components, and associated attributes to ensure that the intended modifications are accurately interpreted and executed. Multimodal grounding is essential to recognize the specific textual or visual document elements referenced by the user. Doc2Command is a multi-task, multimodal Transformer-based model aimed at jointly achieving both these objectives of region of interest segmentation and command generation.

\noindent\textbf{Modeling Doc2Command}: Doc2Command uses a pre-trained Vision Transformer \cite{dosovitskiy2020vit} (ViT) image encoder borrowed from Pix2Struct\cite{lee2023pix2struct} which has been pre-trained with a text decoder for screenshot parsing via masked document image modeling objective. The patch embeddings generated by the encoder serve as input to the pre-trained Pix2Struct decoder and the mask transformer.

\noindent\textbf{Edit Command generation}: We strategically render the input text request as a text box element on the top of the document image. This approach allows for a more flexible integration of linguistic and visual inputs that can be processed jointly by the image encoder. Instead of scaling the input image to a pre-defined resolution, we adjust the scaling factor to maximize the number of fixed-size patches that can fit the image encoder's sequence length. This makes the model more robust against extreme aspect ratios of document images. Each patch is flattened to obtain a vector of pixels and then fed into the image encoder to generate patch encoding. The patch embeddings generated by the encoder serve as input to the text decoder, which auto-regressively generates a sequence of tokens representing the command text specified as: \textit{ACTION(<Component>, 
% <Attribute>, 
<Initial State>, <Final State>)}, containing the action, its associated components, attributes, initial and final states. More details in Sec.~\ref{sec:methodology_doc2command}.

\noindent\textbf{Multimodal Grounding}: We approach the detection of bounding boxes through the lens of a semantic segmentation task. Given the bounding boxes for the region of interest and the rendered user request, we create ground truth segmentation maps with three classes: (1) the Region of Interest, (2) the rendered user request text, and (3) the remaining document. We utilize a DETR-style transformer \cite{detr} for masked attention modeling. A set of $K$ learnable class embeddings ($K = 3$ for our model) is initialized randomly and assigned to a single semantic class. It is used to generate the class mask. The mask-transformer processes the class embeddings jointly with patch encoding and generates $K$ masks by computing the scalar product between L2-normalized patch embeddings with class embeddings output by the decoder. The set of class masks is reshaped into a 2D mask and bilinearly upsampled to the image size to obtain a feature map, followed by a softmax and layer normalization to obtain pixel-wise class scores, forming the final masked segmentation maps that are softly exclusive to each other.
At inference, the segmented area is converted into a bounding box by considering points within a $95\%$ radius of the centroid of the mask. The contours of the largest contiguous object are then used to determine the coordinates of the bounding box, which is denoted by \( (x, y, h, w) \). Here, \( (x, y) \) is the top-left coordinate of the bounding box, \( h \) and \( w \) are height and width, respectively. More details in Sec.~\ref{sec:methodology_grounding}.

\noindent\textbf{Training Doc2Command}: The text decoder is fine-tuned to generate the command text, while the mask transformer is fine-tuned for segmentation. The multitask setup employs a combined weighted loss given by $\mathcal{L}_{\text{total}} = \lambda_{\text{text}} \cdot \mathcal{L}_{\text{text}} + \lambda_{\text{seg}} \cdot \mathcal{L}_{\text{seg}}$. The segmentation loss $\mathcal{L}_{\text{seg}}$ is itself a sum of focal loss \cite{lin2017focal} and dice loss \cite{sudre2017dice}.

% \begin{figure*}
%     \centering
%     \scalebox{1}{
%     \includegraphics[width=0.75\textwidth]{DocEdit-v2 Screenshot.pdf}}
%     \caption{(a) Given a user request, \texttt{DocEdit-v2} uses Doc2Command to ground the edit location in the document image and generate document edit commands, (b) transforms the edit commands into LMM-specific editing instructions using Command Reformulation prompting, and (c) uses GPT-4V to generate edited document HTML.}
% \end{figure*}

\subsection{Command Reformulation Prompting}

%%% Add prompt, input and output and cut down on fluff.

Doc2Command is trained on the command generation task from DocEdit dataset \cite{mathur2023docedit}, which is geared towards generating software-specific commands. Consequently, the generated edit commands are sub-optimal to be used as editing instructions for generalist LMMs
(see examples in Fig.~\ref{fig:eg1}-\ref{fig:eg8}). Additionally, the generated commands may underspecify the actions, components, and associated attributes needed to faithfully produce the final edit due to ambiguities in the user request. Hence, there is a need to reformulate the generated edit commands to perfectly align with the requisite format of the prompt instructions expected by generalist multimodal generation models like GPT-4V and Gemini. We address this limitation by introducing Command Reformulation that leverages in-context learning of Large Language Models (LLMs) to revise the edit commands generated by the Doc2Command module. Fig.~\ref{fig:prompt2} in the Appendix shows the prompt template comprising of the original user request and the edit command from Doc2Command used with an LLM for this purpose. The output from the LLM is an edit instruction customized for LMM-based document editing. Fig. \ref{fig:both} represents two qualitative examples demonstrating command reformulation and the associated impact on the edited document.

\subsection{Generative Document Editing}

\noindent\textbf{HTML+CSS as Document Representations}: Structured textual representations, such as Hypertext Markup Language (HTML) and Cascading Style Sheets (CSS), present notable advantages in alleviating the challenges associated with generative methods in document editing. Firstly, HTML provides a hierarchical structure that inherently captures the organization and relationships among document elements, facilitating the preservation of structural information. This hierarchical representation enables precise manipulation and control over the layout and arrangement of content, which is essential for maintaining document coherence during the editing process. Secondly, CSS decouples content from presentation, offering a systematic approach to capture stylistic attributes such as fonts, colors, and layouts. This separation of content and style allows for greater flexibility in rendering documents while preserving their underlying structure. Hence, we conceptualize document editing as a text generation task by expressing the document as an HTML+CSS rendering.

\noindent\textbf{Generating HTML+CSS Data}: We employ generative large multimodal models (LMMs), specifically GPT-4V and Gemini, to convert both the input as well as ground truth document images into a closely replicated HTML and CSS rendering via constraint-driven prompt engineering. Our experimental setup imposes strict constraints on the generated HTML documents to ensure standardization across class names, adequate utilization of flexbox for layouts, higher preference for embedded CSS, and replacement of visual media with placeholders. Maintaining consistency and coherence across the generated HTML+CSS facilitates fair evaluation.

\noindent\textbf{LMM Prompting}: We utilize multimodal prompting of GPT-4V and Gemini 
by incorporating the set of marks \cite{yang2023setofmark} for the grounded RoI bounding boxes extracted by Doc2Command and the edit instruction produced in the Command Reformulation step. Such multimodal prompting guides LMMs to closely adhere to the provided commands while paying special attention to the visual cues specified by the bounding box in the document image. This ensures that the generated edits accurately reflect the intended modifications.

\section{Document Editing Evaluation}
We perform system output evaluation as follows:

\noindent\textbf{Automated Metrics}: Apart from the document metrics reported by \citet{mathur2023docedit} for command text generation (Exact Match, ROUGE-L, Word Overlap F1, Action and Component Accuracy $\%$) and RoI bounding box prediction (Top-1 accuracy $\%$), we adapt two novel metrics, specific to HTML document editing: 

\textbf{(1) DOM Tree Edit Distance} -- Document Object Model (DOM) tree represents the hierarchical structure of the HTML document. Comparing the DOM tree of two HTML documents yields information about their structural differences.  We utilize the Zhang-Shasha algorithm \cite{Zhang1989SimpleFA} to calculate the edit distance between the generated and ground truth DOM trees.

\textbf{(2) CSS IoU}: Cascading Style Sheets (CSS) deal with the presentation of HTML documents and dictates how they would be rendered. In recreating document images into HTML pages, CSS in the form of property-value pairs of different attributes controls the formatting, style and layout of the rendered HTML document. Sets of property-value pairs from inline CSS and internal CSS selectors are obtained, and the Intersection over Union (IoU) is calculated over these sets to evaluate the similarity between the styles of the edited and ground truth documents. We also evaluate parallel HTML documents using ROUGE-L and Word Overlap F1, applied to the entire document.

\noindent\textbf{Human Evaluation}: Every edited document HTML is evaluated by three human evaluators on our three proposed metrics: 
\textbf{(1) Style Replication} assesses whether the styles of the original document are preserved, \textbf{(2) Content Replication} evaluates if the textual content of the region of non-interest in the original document HTML is conserved, \textbf{(3) Edit Correctness}: judges whether the user's editing intent has been faithfully fulfilled. Each of these metrics yields a binary score, which is averaged across evaluators and then summed to compute a unified score for each document.

\begin{figure}[h]
    \centering
    \begin{subfigure}[b]{0.4\textwidth}
        \includegraphics[width=\textwidth]{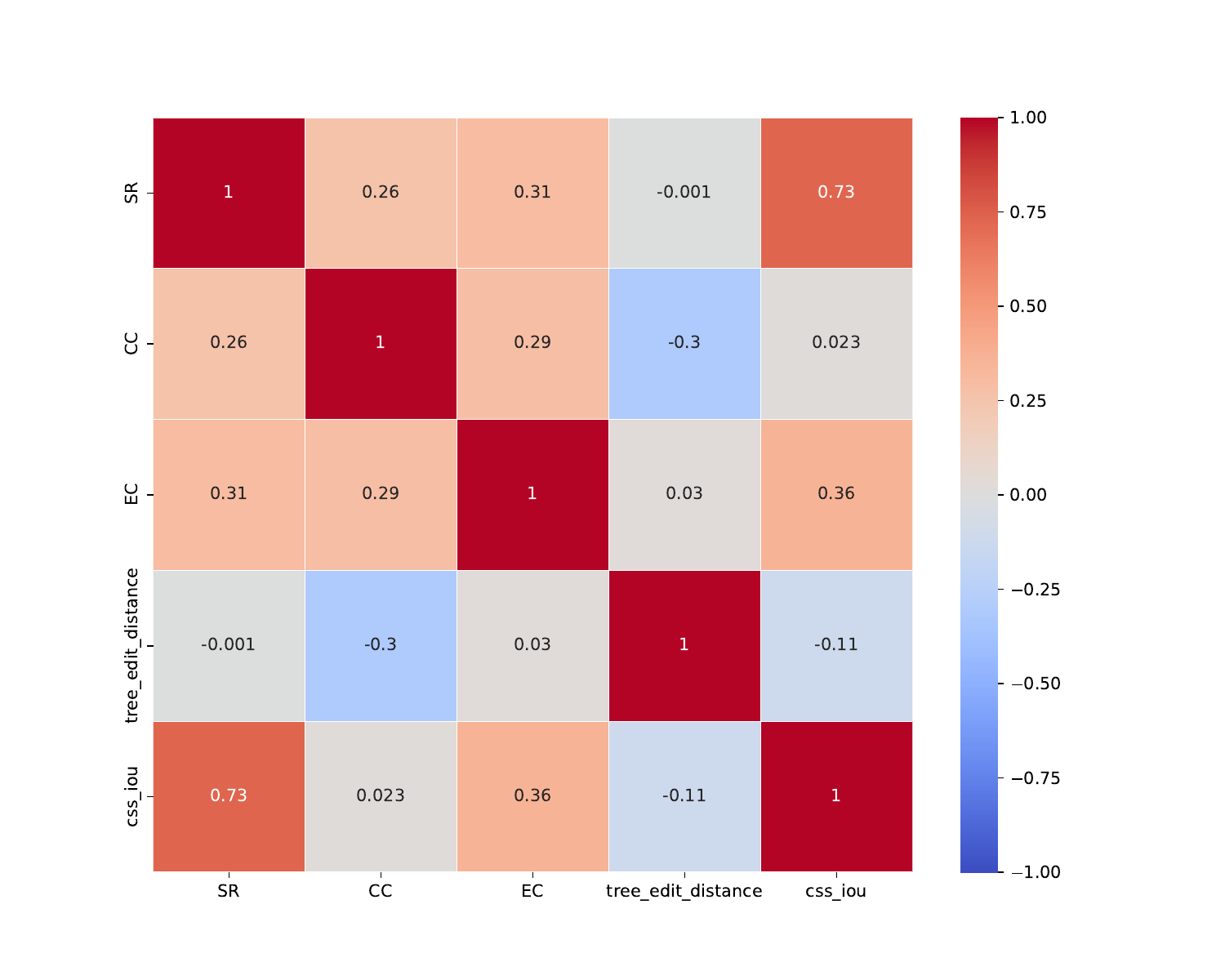}
        \caption{Correlation Heatmap: Tree Edit Distance and CSS IoU with Human Evaluation Metrics (Style Replication, Content Replication, Edit Correctness).}
        \label{fig:heatmap1}
    \end{subfigure}
    \hfill
    \begin{subfigure}[b]{0.37\textwidth}
        \includegraphics[width=\textwidth]{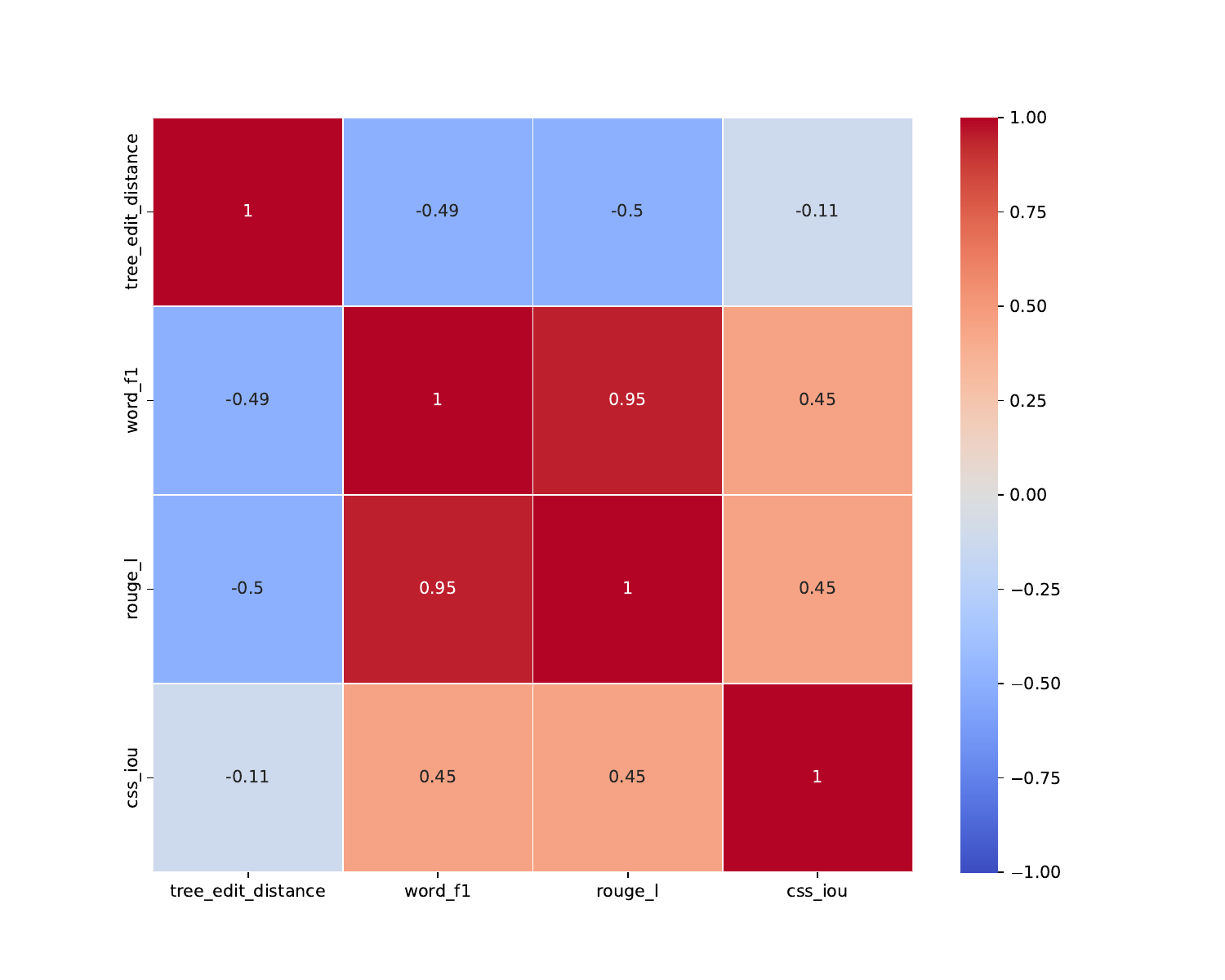}
        \caption{Correlation Heatmap: Tree Edit Distance and CSS IoU with Automated Metrics (Word F1, ROUGE-L).}
        \label{fig:heatmap2}
    \end{subfigure}
    \caption{Correlation Heatmaps for Tree Edit Distance and CSS IoU with Human and Automated Evaluation Metrics.}
    \label{fig:heatmaps}
\end{figure}

The heatmaps in Figure \ref{fig:heatmaps} compare our proposed metrics, Tree Edit Distance and CSS IoU, against both human and automated evaluation metrics. In human evaluations, CSS IoU shows a strong correlation (0.73) with Style Replication, highlighting its sensitivity to visual presentation. Tree Edit Distance, however, compares HTML document structures, which do not directly relate with any human evaluation parameters, showing no significant correlation results. These results demonstrate the importance of human evaluation supplementary to metric based evaluation. When compared to automated metrics like Word Overlap F1 and ROUGE-L, Tree Edit Distance moderately correlates negatively (-0.49, -0.50), as expected, since a higher tree edit distance reflects dissimilar documents. Tree Edit Distance, has a moderate positive correlation with both metrics (0.45), implying that presentation style partially influences text-based overlap. 

\section{Experimental Settings}
\label{sec:experiments}

\subsection{Data}
\label{sec:datasets}
We utilize the DocEdit-PDF dataset, introduced by \citet{mathur2023docedit}. The dataset comprises pairs of 17,808 document images, with corresponding user edit requests and ground truth edit commands. Our experiments are conducted on the default data split provided in the official dataset release, wherein the data is partitioned into training, testing, and validation sets in an 8:2:1 ratio. All reported results are based on the test set. The license for the dataset can be found \href{https://github.com/adobe-research/DocEdit-Dataset/blob/main/LICENSE}{here}.

\subsection{Implementation Details}
\textbf{Doc2Command}
Our experiments utilized the Adafactor optimization algorithm with a learning rate of $3 \times 10^{-5}$ and weight decay set to $1 \times 10^{-5}$. The training process spanned 30 epochs with a batch size of 1. The input data was organized into patches of size 16, limiting the maximum number of patches to 1024. The learning rate was scheduled using a cosine scheduler with a warm-up period equivalent to 10\% of the iterations within each epoch. For loss computation, we introduced loss weighing factors $\lambda_{\text{text}} = 0.3$ and $\lambda_{\text{seg}} = 1.5$. The sigmoid focal loss was utilized for segmentation with parameters $\alpha = 0.25$ and $\gamma = 2$. Additionally, the decoder included a dropout rate of 0.1.

\noindent\textbf{Command Reformulation and Document Editing}: We use \texttt{gpt-4} \cite{OpenAI2023GPT4TR} and \texttt{gemini-pro} \cite{team2023gemini} for command reformulation, and \texttt{gpt-4-vision-preview}/\texttt{gemini-pro-vision} for document editing. 
% The models are accessed using the OpenAI API \footnote{\url{https://platform.openai.com/}}.
We set the temperature parameter to 0 to ensure deterministic and reproducible experiments and use the default value for all other parameters. The visual grounding and command grounding are obtained by inferring Doc2Command on the test set. The maximum token count for the output is set as 4000.

One limitation of using HTML as a medium to express document edits is that the ground truth post-edit documents only exist as document images, with bounding boxes to indicate edited regions. Therefore, we generate HTML replications of the ground truth post-edit documents using LMMs. To ensure consistency, we use the same prompt details for image-to-HTML conversion as the document editing experiments. Additionally, we prompt the model to pay special attention to the style and content in the bounding box while recreating the document image as an HTML document. We perform human evaluation of the ground truth post-edit HTML documents by comparing them to ground truth images as described in the Metrics subsection and find that style replication score and content replication score are 75.23\% and 92.3\% (GPT-4V), and 70.14\% and 87\% (Gemini) respectively, with a Cohen's Kappa score $\ge 0.84$ across evaluators and tasks. More implementation details on the metrics (Sec.~\ref{sec:eval_metrics}), computational resources (Sec.~\ref{sec:compute}, and human evaluations (Sec.~\ref{sec:eval_instructions}) are in the Appendix.

\section{Baselines}

\begin{table*} %[h]
    \centering
    \resizebox{1.6\columnwidth}{!}{%
    \begin{tabular}{lcccccc}
                \toprule
                \textbf{System} & \textbf{EM (\%)} & \textbf{Word Overlap F1} & \textbf{ROUGE-L} & \textbf{Action (\%)} & \textbf{Component (\%)} \\
                \midrule
            Generator-Extractor & 6.6 & 0.25 & 0.22 & 36.7 & 8.5 \\
            GPT2 \cite{gpt2}& 11.6 & 0.76 & 0.76 & 79.7 & 27.2 \\
            BART \cite{lewis-etal-2020-bart}& 19.7 & 0.78 & 0.76 & 81.2 & 29.5 \\
            T5 \cite{2020t5}& 20.4 & 0.79 & 0.76 & 81.4 & 29.8 \\
            BERT2GPT2 & 7.3 & 0.37 & 0.39 & 45.2 & 9.2 \\
            LayoutLMv3-GPT2 & 8.7 & 0.39 & 0.40 & 47.6 & 10.3 \\
            CLIPCap \cite{clipcap} & 8.5 & 0.25 & 0.27 & 44.5 & 9.34 \\
            DiTCap \cite{dit} & 23.6 & 0.81 & 0.80 & 82.5 & 25.5 \\
            Multimodal Transformer \cite{multimodaltransformer} & 31.6 & 0.82 & 0.83 & 83.1 & 32.4 \\
            DocEditor \cite{mathur2023docedit} & \cellcolor[HTML]{FFCCC9}37.6 & \cellcolor[HTML]{FFCCC9}0.87 & \cellcolor[HTML]{FFCCC9}0.83 & \cellcolor[HTML]{FD6864}87.6 & 40.7 \\
            GPT3.5 \cite{gpt3} & 10.1 & 0.77 & 0.77 & 75.93 & 73.37 \\
            GPT4 \cite{OpenAI2023GPT4TR} & 14.3 & 0.78 & 0.78 & 81.57 & \cellcolor[HTML]{FFCCC9}75.03  \\
            \midrule
            \textbf{Doc2Command} & \cellcolor[HTML]{FD6864}39.6 & \cellcolor[HTML]{FD6864}0.87 & \cellcolor[HTML]{FD6864}0.86 & \cellcolor[HTML]{FFCCC9}85.0 & \cellcolor[HTML]{FD6864}86.1 \\
            \bottomrule
            \end{tabular}%
    }
    
    \caption{Results for the command generation task. Doc2Command shows the best performance (see \colorbox[HTML]{FD6864}{Red}).}
    \label{tab:results_command}
\end{table*}

\begin{table} %[h]
    \centering
    \resizebox{0.6\columnwidth}{!}{%
    \begin{tabular}{lc}
            \toprule
            \textbf{System} & \textbf{Top-1 Acc (\%)} \\
            \midrule
            ReSC-Large \cite{resc} & 17.04 \\
            Trans VG \cite{deng2022transvg} & 25.34 \\
            DocEditor \cite{mathur2023docedit} & \cellcolor[HTML]{FFCCC9}36.50 \\
            \midrule
            \textbf{Doc2Command} & \cellcolor[HTML]{FD6864}48.69 \\
            \bottomrule
        \end{tabular}%
    }
    \caption{Results for bounding box detection task. Doc2Command shows the best performance (see \colorbox[HTML]{FD6864}{Red}).}
    \label{tab:results_bbox}
\end{table}

\begin{table*} %[h!]
\centering
\resizebox{1.7\columnwidth}{!}{%
\begin{tabular}{lcccrrrrrrrr}
\hline
\multicolumn{4}{c}{\textbf{Experimental Setting}} & \multicolumn{4}{c}{\textbf{Automated Evaluation}} & \multicolumn{4}{c}{\textbf{Human Evaluation}} \\ \hline
\textbf{Method} & \multicolumn{1}{l}{\textbf{VG}} & \multicolumn{1}{l}{\textbf{CG}} & \multicolumn{1}{l|}{\textbf{CR}} & \multicolumn{1}{l}{\textbf{ROUGE-L}} & \multicolumn{1}{l}{\textbf{Word Overlap F1}} & \multicolumn{1}{l}{\textbf{Tree Edit Distance}} & \multicolumn{1}{l|}{\textbf{CSS IoU}} & \multicolumn{1}{l}{\textbf{SR (\%)}} & \multicolumn{1}{l}{\textbf{EC (\%)}} & \multicolumn{1}{l}{\textbf{CC (\%)}} & \multicolumn{1}{l}{\textbf{Total Score (\%)}} \\ \hline
 GPT-4V Only & {\xmark} & {\xmark} & \multicolumn{1}{c|}{N/A} & 0.406 & 0.451 & 24.13 & \multicolumn{1}{r|}{0.245} & 73.53 & 27.45 & 66.77 & 55.92 \\ \hline
 & {\cmark} & {\xmark} & \multicolumn{1}{c|}{N/A} & 0.410 & 0.460 & 24.02 & \multicolumn{1}{r|}{0.250} & 74.28 & 45.28 & 68.21 & 62.59 \\
 & {\xmark} & {\cmark} & \multicolumn{1}{c|}{{\xmark}} & 0.412 & 0.458 & 23.54 & \multicolumn{1}{r|}{0.247} & 75.02 & 49.32 & 68.22 & 64.19 \\
 & {\xmark} & {\cmark} & \multicolumn{1}{c|}{{\cmark}} & 0.409 & 0.455 & \cellcolor[HTML]{FFCCC9}23.27 & \multicolumn{1}{r|}{0.245} & 74.87 & 51.87 & \cellcolor[HTML]{FFCCC9}69.71 & 65.49 \\
 \multirow{-4}{*}{GPT-4V +}  & {\cmark} & {\cmark} & \multicolumn{1}{c|}{{\xmark}} & \cellcolor[HTML]{FFCCC9}0.416 & \cellcolor[HTML]{FFCCC9}0.461 & 23.72 & \multicolumn{1}{r|}{\cellcolor[HTML]{FFCCC9}0.251} & \cellcolor[HTML]{FFCCC9}75.14 & \cellcolor[HTML]{FFCCC9}55.33 & \cellcolor[HTML]{FD6864}69.89 & \cellcolor[HTML]{FFCCC9}66.79\\ \hline
\textbf{DocEdit-v2} & {\cmark} & {\cmark} & \multicolumn{1}{c|}{{\cmark}} & \cellcolor[HTML]{FD6864}0.417 & \cellcolor[HTML]{FD6864}0.463 & \cellcolor[HTML]{FD6864}23.15 & \multicolumn{1}{r|}{\cellcolor[HTML]{FD6864}0.252} & \cellcolor[HTML]{FD6864}75.31 & \cellcolor[HTML]{FD6864}57.41 & 69.14 & \cellcolor[HTML]{FD6864}67.28 \\ \hline
\end{tabular}%
}
\caption{Results and ablations for end-to-end document editing task using GPT-4V as the base LMM. Here, VG = Visual Grounding, CG = Command Generation, and CR = Command Reformulation. \colorbox[HTML]{FD6864}{Red} represents best performance.}
\label{tab:editing_res}
\end{table*}

\begin{table*}[h]
\centering
\resizebox{1.7\columnwidth}{!}{%
\begin{tabular}{lcccrrrrrrrr}
\hline
\multicolumn{4}{c}{\textbf{Experimental Setting}} & \multicolumn{4}{c}{\textbf{Automated Evaluation}} & \multicolumn{4}{c}{\textbf{Human Evaluation}} \\ \hline
\textbf{Method} & \multicolumn{1}{l}{\textbf{VG}} & \multicolumn{1}{l}{\textbf{CG}} & \multicolumn{1}{l|}{\textbf{CR}} & \multicolumn{1}{l}{\textbf{ROUGE-L}} & \multicolumn{1}{l}{\textbf{Word Overlap F1}} & \multicolumn{1}{l}{\textbf{Tree Edit Distance}} & \multicolumn{1}{l|}{\textbf{CSS IoU}} & \multicolumn{1}{l}{\textbf{SR (\%)}} & \multicolumn{1}{l}{\textbf{EC (\%)}} & \multicolumn{1}{l}{\textbf{CC (\%)}} & \multicolumn{1}{l}{\textbf{Total Score (\%)}} \\ \hline
 Gemini Only& {\xmark} & {\xmark} & \multicolumn{1}{c|}{N/A} & 0.438& 0.542& 62.95& \multicolumn{1}{r|}{0.333} & 59.64& 15.79& 61.41& 45.61\\ \hline
 & {\cmark} & {\xmark} & \multicolumn{1}{c|}{N/A} & 0.447& 0.551& 54.63& \multicolumn{1}{r|}{0.332} & 60.12& 39.22& 65.02& 54.79\\
 & {\xmark} & {\cmark} & \multicolumn{1}{c|}{{\xmark}} & \cellcolor[HTML]{FFCCC9}0.451& 0.544& 65.06& \multicolumn{1}{r|}{0.334} & 61.92& 37.65& 64.28& 54.62\\
 & {\xmark} & {\cmark} & \multicolumn{1}{c|}{{\cmark}} & 0.417& 0.510& \cellcolor[HTML]{FFCCC9} 53.89& \multicolumn{1}{r|}{0.341} & 62.52& 40.44& \cellcolor[HTML]{FFCCC9}67.11& 56.69\\
 \multirow{-4}{*}{Gemini +}& {\cmark} & {\cmark} & \multicolumn{1}{c|}{{\xmark}} & 0.437& \cellcolor[HTML]{FFCCC9} 0.554& 55.41& \multicolumn{1}{r|}{\cellcolor[HTML]{FFCCC9}0.342} & \cellcolor[HTML]{FD6864}64.12& \cellcolor[HTML]{FFCCC9}41.35& 66.96& \cellcolor[HTML]{FFCCC9} 57.48\\ \hline
\textbf{DocEdit-v2} & {\cmark} & {\cmark} & \multicolumn{1}{c|}{{\cmark}} & \cellcolor[HTML]{FD6864}0.454& \cellcolor[HTML]{FD6864}0.557& \cellcolor[HTML]{FD6864}52.24& \multicolumn{1}{r|}{\cellcolor[HTML]{FD6864}0.367} & \cellcolor[HTML]{FFCCC9}63.16& \cellcolor[HTML]{FD6864}44.73& \cellcolor[HTML]{FD6864}68.42& \cellcolor[HTML]{FD6864}58.77\\ \hline
\end{tabular}%
}
\caption{Results and ablations for end-to-end document editing task using Gemini as the base LMM. Here, VG = Visual Grounding, CG = Command Generation, and CR = Command Reformulation. \colorbox[HTML]{FD6864}{Red} represents best performance.}
\label{tab:editing_res2}
\end{table*}

\label{sec:baselines}
\textbf{Command Grounding Baselines:}
We investigate several command generation baselines to establish performance benchmarks. Initially, we employ Seq2Seq text-only models, including GPT2 \cite{gpt2}, BART \cite{lewis-etal-2020-bart}, and T5 \cite{2020t5}, which exclusively process user text descriptions. Subsequently, we explore the Generator-Extractor paradigm, integrating BERT \cite{devlin2019bert} and DETR \cite{detr} with autoregressive decoding for command generation. Additionally, we examine Transformer Encoder-Decoder architectures, such as LayoutLMv3-GPT2 and BERT2GPT2 \cite{huang2022layoutlmv3}, which combine GPT2 decoders with LayoutLMv3 and BERT encoders, respectively. Furthermore, we investigate Prefix Encoding \cite{clipcap}, utilizing learned representations from pre-trained encoders like CLIP \cite{clip} and DiT \cite{dit} as a prefix to the GPT2 decoder network. Additionally, we consider the Multimodal Transformer\cite{multimodaltransformer}, which incorporates multimodal input from user descriptions, visual objects, and document text to generate commands. Moreover, we explore DocEditor \cite{mathur2023docedit}, a task-specific baseline employing a Transformer-based multimodal model that decomposes document images into OCR content and object boxes, utilizing multimodal transformers to generate commands. Finally, we compare against GPT3.5 \cite{gpt3} and GPT4 \cite{OpenAI2023GPT4TR}, employing in-context learning by providing three examples of each command type as context to the model for evaluation.\textbf{Visual Grounding Baselines:}
We consider several baselines for bounding box detection in the context of visual grounding for document editing. Firstly, ReSC-Large \cite{resc} presents a method for direct coordinates regression in the Region of Interest (RoI) bounding box prediction task. Similarly, TransVG \cite{deng2022transvg} offers an alternative approach for direct coordinates regression in RoI bounding box prediction. Additionally, we investigate DocEditor \cite{mathur2023docedit}, which employs a comprehensive methodology. DocEditor initially encodes the document image by extracting text through Optical Character Recognition (OCR) and utilizes object detection to capture visual features. Subsequently, transformer-encoded features are fed into a Gated Relational Graph Convolutional Network (R-GCN) to generate a layout graph-aware representation. This representation is then leveraged downstream to perform bounding box regression, facilitating accurate localization of document elements.
\textbf{Document Editing Baselines:}
Certain experimental configurations are employed to investigate the effectiveness of command reformulation and multimodal grounding in harnessing the capabilities of GPT-4V and Gemini as document editing tools. Specifically, visual grounding, command grounding, and command reformulation are selectively excluded from our experiments. In this context, command grounding is supplanted by the unstructured user request, while visual grounding is eliminated by presenting the original document image as the input, thus eliminating the need for explicit visual cues (rendered bounding boxes). Moreover, command reformulation is eliminated by directly utilizing the command generated by the Doc2Command model. Notably, the absence of command grounding renders command reformulation inapplicable (N/A), as the reformulation process relies on refining commands derived from grounded contexts.

\section{Results}

\noindent\textbf{Edit Request Grounding}: Table \ref{tab:results_command} shows the performance of \texttt{DocEdit-v2} against contemporary baselines for command generation tasks. \texttt{DocEdit-v2} achieves an impressive $86.1\%$ accuracy in recognizing document components 
%and $86$
, outperforming the previous state-of-the-art (SoTA) by $~10.7\%$. We see consistent gains for the exact match accuracy and ROUGE-L score, although comparable performance to SOTA across action accuracy $(\%)$ and word overlap F1. We show significant improvement in component accuracy $(\%)$ over the previous task specific SoTA, $45\%$ points.  We attribute this notable improvement to the Doc2Command module, which can effectively comprehend natural language requests and ground them into complex document structures and layouts. Table \ref{tab:results_bbox} shows that Doc2Command yields remarkable enhancements in the bounding box detection task with a Top-1 accuracy of $48.69\%$, surpassing the previous SoTA performance by $12.19\%$, which further signifies our system's effectiveness in accurately grounding edit requests to document images.

\noindent\textbf{Generative Document Editing}: Table \ref{tab:editing_res} and \ref{tab:editing_res2} shows the results for end-to-end document editing task with GPT-4V and Gemini as the base LMMs respectively. We observe that Doc2Command and Command Reformulation prompting are critical components as removing either severely deteriorates performance across automated and human evaluations. We observe \verb|~|2-3 \% decline in Edit Correction when command reformulation prompting is removed (in both settings: with or without visual grounding) . Visual grounding assists by localising the edit region, which can be demonstrated by an improvement of \verb|~|$18-23\%$ when GPT-4V is prompted with visual grounding.

Significant performance gains across Tree Edit Distance and CSS IoU indicate the ability of GPT-4V and Gemini to consistently recreate non-RoI parts of the document, proving the effectiveness of editing HTML and CSS directly. The experiment setting with no multimodal grounding performs worst, while multimodal grounding with command reformulation improves editing correctness (EC) by 29.96\%(GPT-4V)/28.94\%(Gemini) and overall human evaluation score by 11.36\%(GPT-4V)/13.16\%(Gemini).

Fig \ref{fig:eg1}-\ref{fig:eg9} show qualitative examples of document editing by \texttt{DocEdit-v2} for diverse edit requests such as spatial alignment, component placement, text paraphrasing and applying special effects which involve manipulating and rendering different document elements such as text, tables, figures and lists.

\section{Conclusion}
We introduce the \texttt{DocEdit-v2} framework for end-to-end document editing. DocEdit-v2 draws on Doc2Command, a multi-task multimodal model that visually localizes user requests in the document image and generates edit commands, which are further refined using Command Reformulation prompting. \texttt{DocEdit-v2} uses LMMs multimodal prompting with request grounding and edit instructions to perform generative editing of the HTML+CSS structure of documents, showcasing remarkable performance improvements across editing accuracy, command generation, and RoI detection. Future work will aim to enhance the framework's adaptability to diverse document types, including multi-page documents.

\section{Ethics Statement}
We utilize the publicly available DocEdit-PDF corpus for this research without introducing new annotations. We use publicly available API-accessible LMMs and LLMs for our experiments. The identity of the human evaluators is confidential and private. We do not utilize any PII at any step in our experiments. The intended applications of our work are strictly limited to the document editing domain. We refer users to relevant works by \citep{risks1, risks2, risks3} to understand risks and some mitigation strategies for LLM safety.

\section{Limitations}
\begin{enumerate}
    \item \textbf{Document Recreation} The DocEdit Corpus \cite{mathur2023docedit} has documents only as document images. Pixel level manipulation of text-dense image is a challenge, hence we prompt LMMs to produce faithful HTML+CSS recreations. The HTML+CSS documents are close but not identical to the original document images.
     
    \item \textbf{Visual Elements} \texttt{DocEdit-v2} is constrained with generating edited documents as HTML+CSS documents. Complex visual elements such as charts and figures cannot be generated using simple HTML and CSS. Moreover, the transformer backbone used in Doc2Command is pre-trained primrarily on text-dominant document images and has limitations in grounding requests manipulating these visual elements.
    
    \item \textbf{Large Multimodal Models} Our work utilizes API-accessible Large Multimodal Models (LMMs). Model APIs have an associated cost which depends on the token count in the request and model response, image resolution and dimensions. These API based models are also prone to performance fluctuations.
    
\end{enumerate}
\bibliography{custom}

\appendix
\section{Appendix}
\label{sec:appendix}

% \subsection{Related Work}

% Past works in the domain of language-guided image editing have predominantly centered on natural image datasets \cite{image_dataset1, image_dataset2}, overlooking the distinctive characteristics of documents, which typically exhibit text-rich content alongside a diverse array of structured elements arranged in various layouts. These datasets often lack representations of localized edits and indirect edit references, crucial facets for effective document editing. Notably, contemporary GAN-based \cite{gan1, gan2, gan3, gan4, gan5} and diffusion methods \cite{diffusion1, diffusion2, diffusion3, diffusion4, diffusion5} have gained traction for natural image manipulation tasks due to their capacity for end-to-end pixel-level image synthesis. However, their applicability to digital documents, characterized by rich textual content and complex layouts, remains limited. These techniques are ill-equipped to grasp the spatial and semantic intricacies inherent in embedded textual components within documents. Consequently, prior endeavors in language-guided document editing have primarily pivoted towards multimodal grounding of edit requests through textual and visual cues into actionable commands and visual localization \cite{mathur2023docedit}. Despite these efforts, the absence of efficient generative frameworks tailored for document image editing remains a significant challenge in this domain. 

\subsection{Examples}
\begin{figure*}[h!]
    \centering
    \begin{subfigure}{0.6\columnwidth}
        \includegraphics[width=\linewidth]{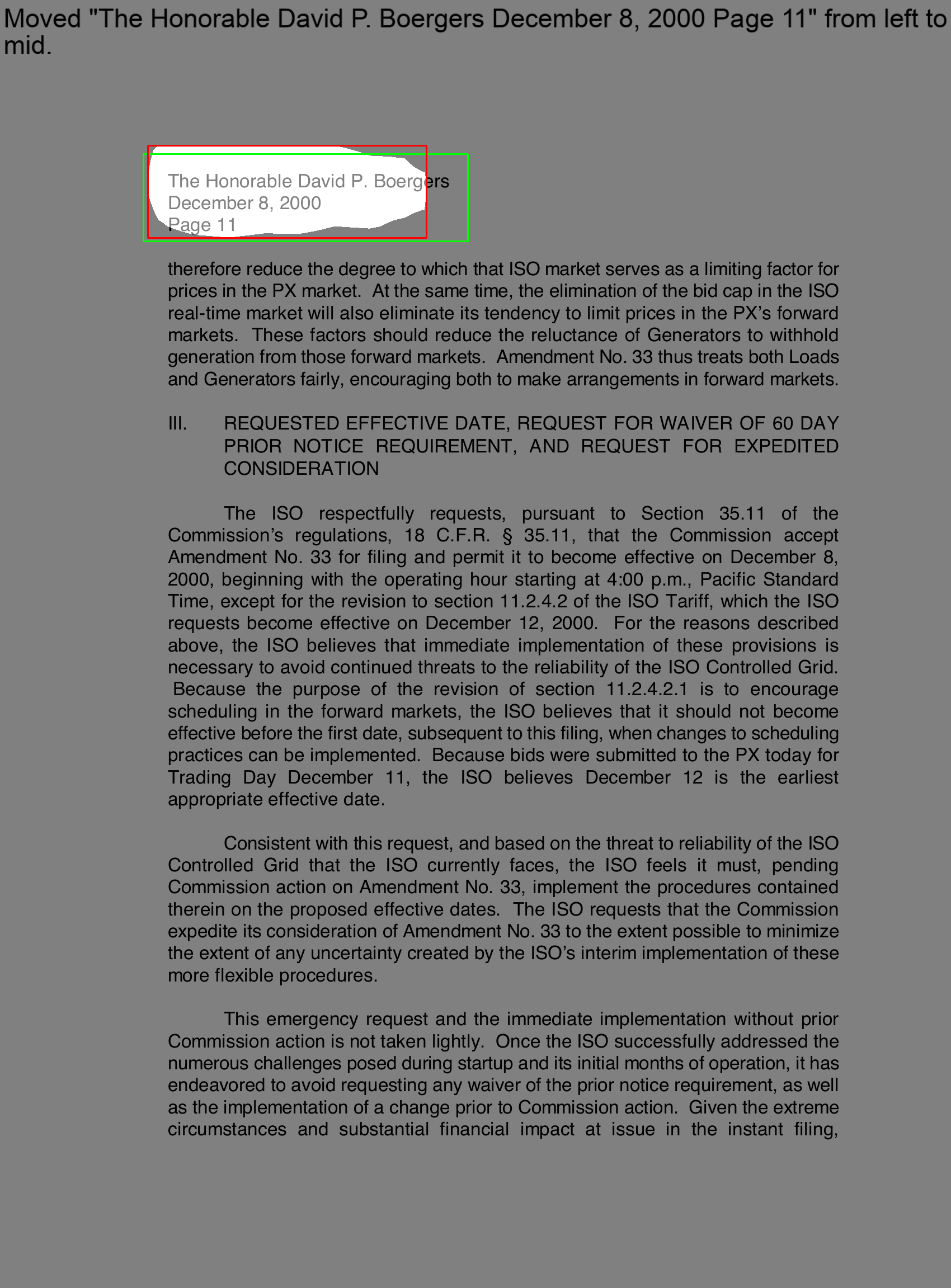}
        \caption{Bounding Box with high IOU: capability to read and recognise text from request in the document.}
    \end{subfigure}\hfill
    \begin{subfigure}{0.6\columnwidth}
        \includegraphics[width=\linewidth]{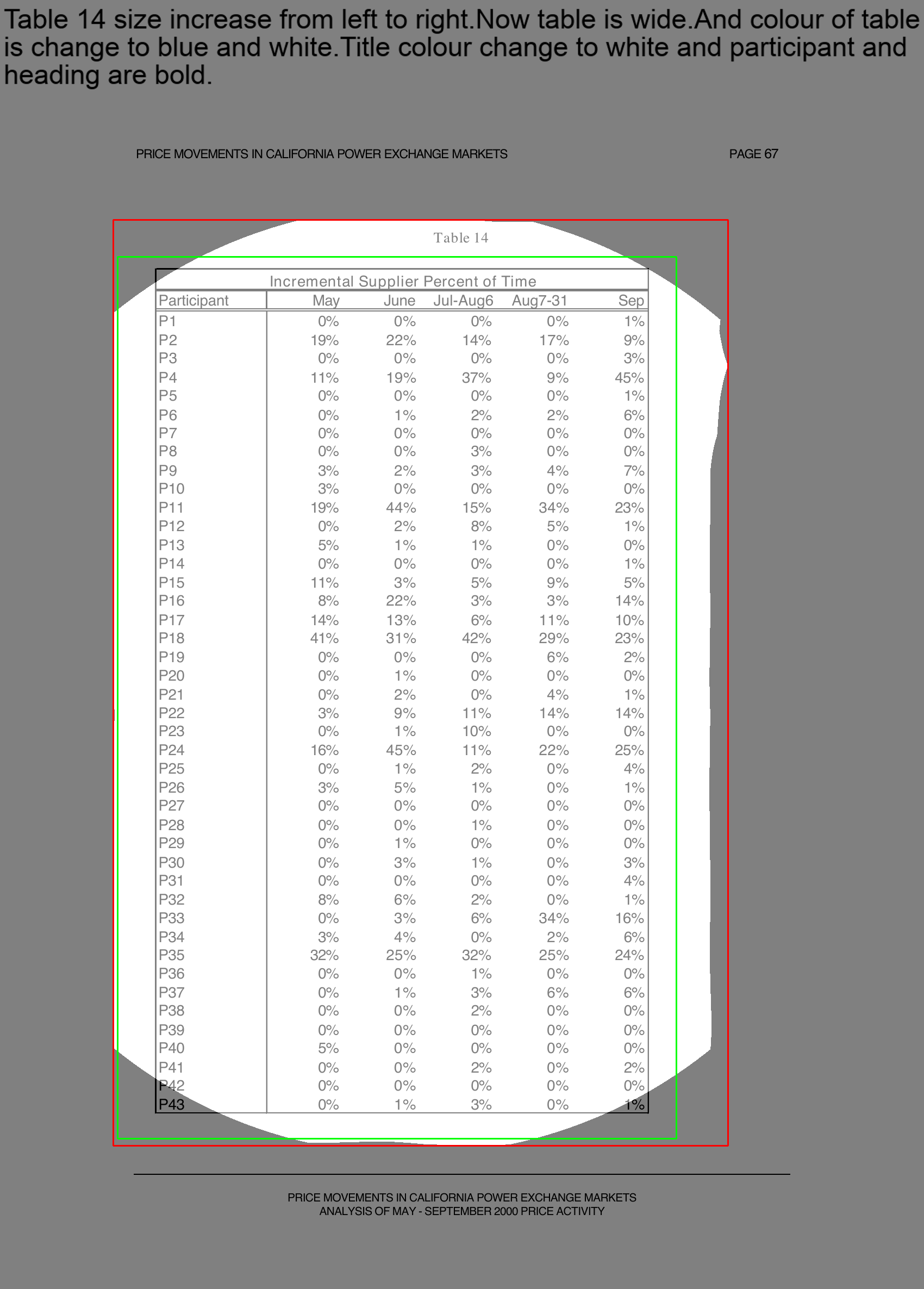}
        \caption{Bounding Box with high IOU: capability to recognise elements such as tables.}
    \end{subfigure}
    \hfill
    \begin{subfigure}{0.6\columnwidth}
        \includegraphics[width=\linewidth]{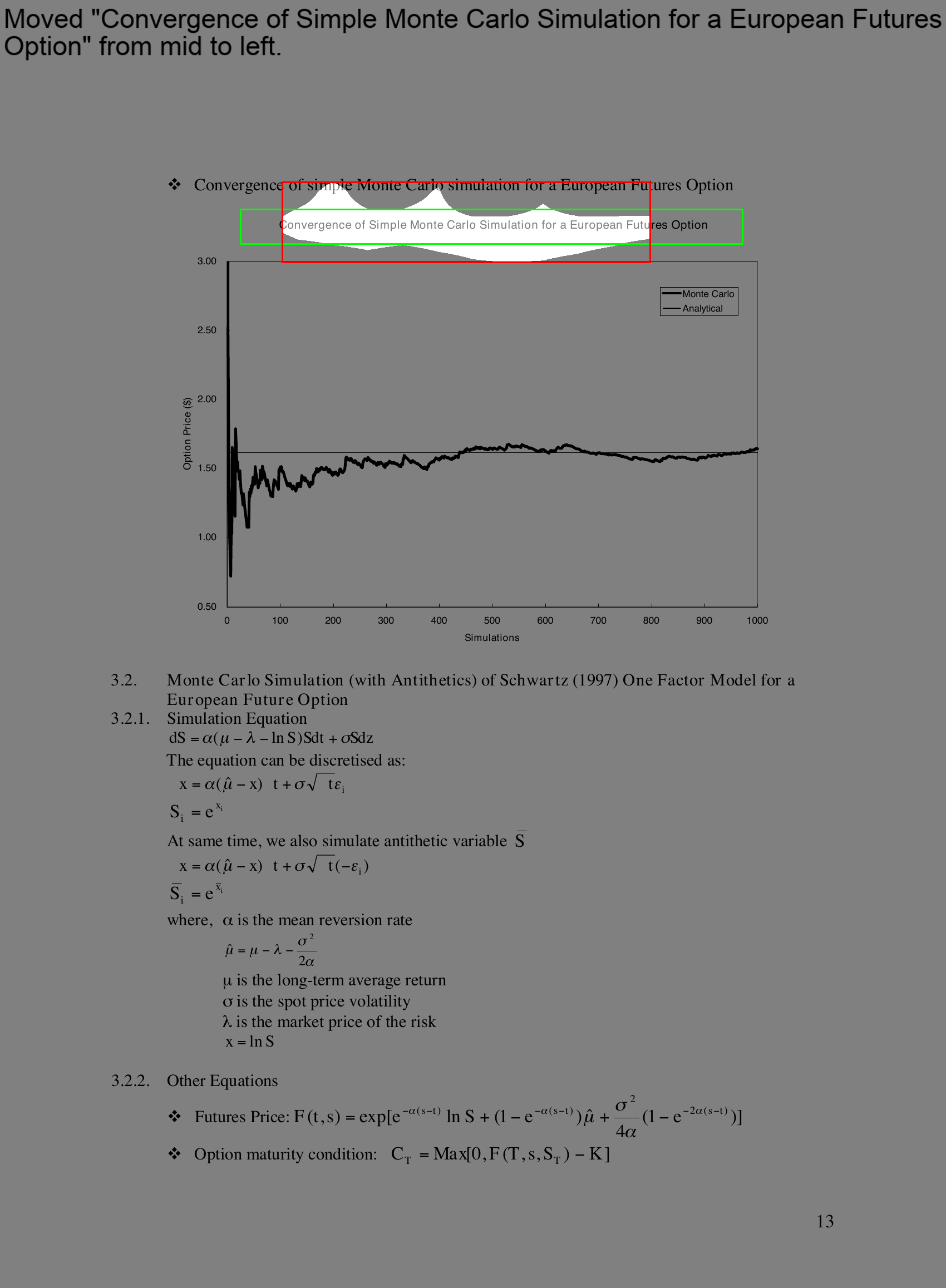}
        \caption{Bounding Box with high IOU: When given two elements with the same text, capability to localize based on position reference.}
    \end{subfigure}

    % \medskip

    \begin{subfigure}{0.6\columnwidth}
        \includegraphics[width=\linewidth]{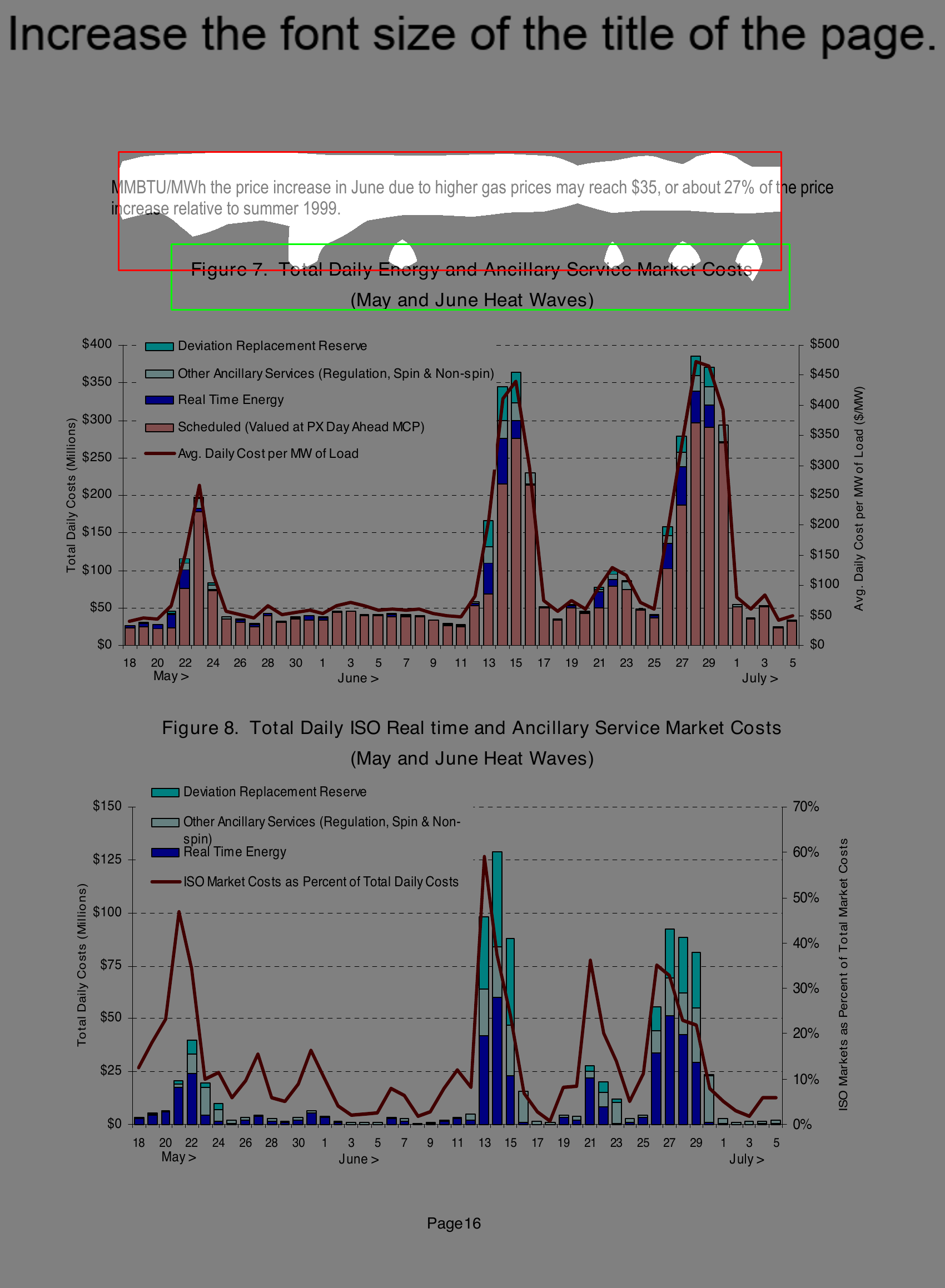}
        \caption{Bounding Box with low IOU: Ambiguity in the page's title.}
    \end{subfigure}\hfill
    \begin{subfigure}{0.6\columnwidth}
        \includegraphics[width=\linewidth]{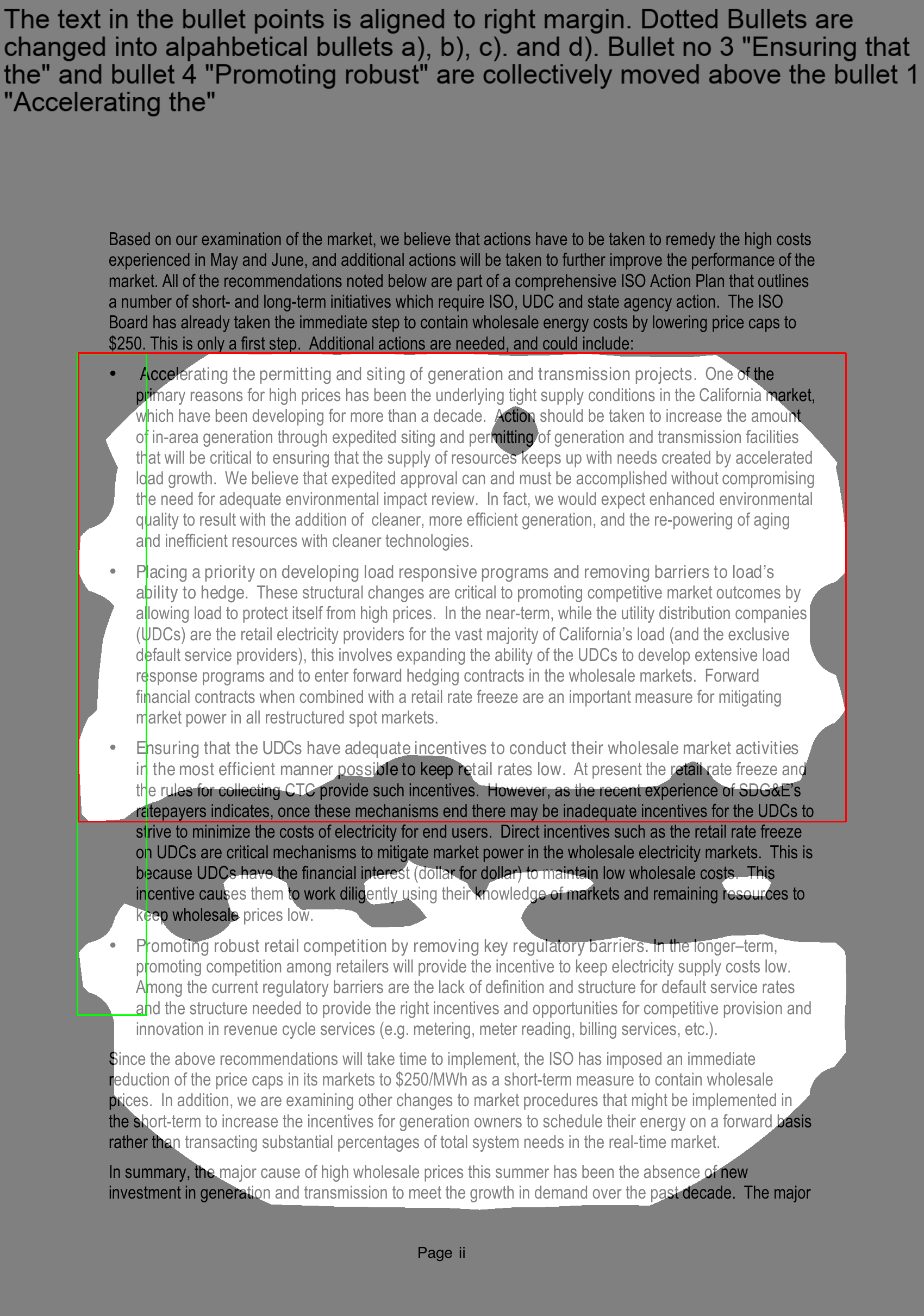}
        \caption{Bounding Box with low IOU: Mask highlights the points that have been bulleted but not the bullets exclusively.}
    \end{subfigure}
    \hfill
    \begin{subfigure}{0.6\columnwidth}
        \includegraphics[width=\linewidth]{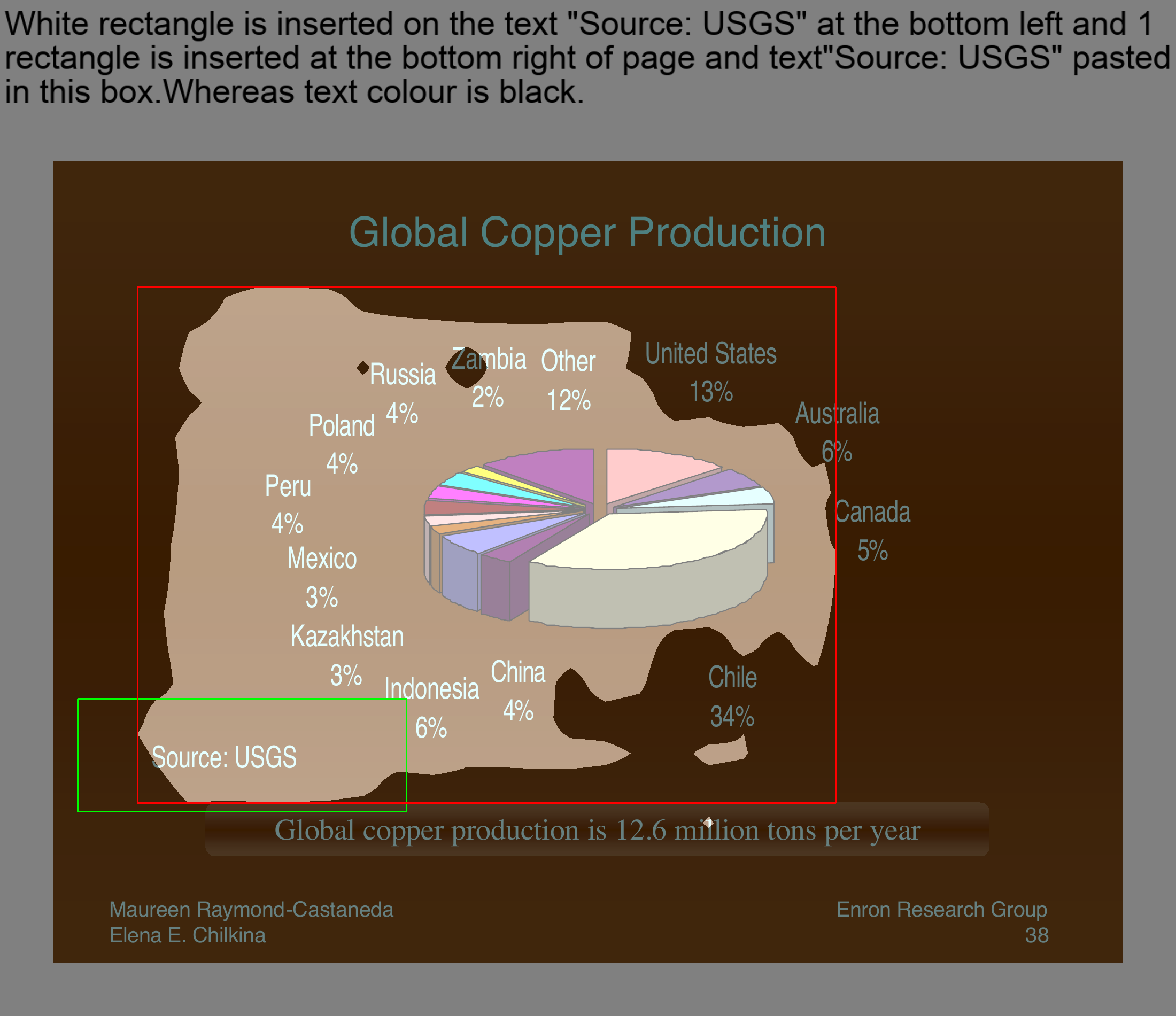}
        \caption{Bounding Box with low IOU: edit request involves text in visual elements}
    \end{subfigure}

    \caption{Examples of segmentation outputs and bounding boxes. The bright white areas represent segmentation outputs. Green boxes represent ground truth bounding boxes, and red boxes represent the inferred bounding boxes.}
    \label{fig:four_images}
\end{figure*}

Fig. \ref{fig:four_images} represents 6 examples of our model's performance on the test set. Subfigures (a), (b), and (c) represent correctly inferred examples, and (d), (e), and (f) represent incorrectly inferred examples. With each example, the figure explains the capability or limitation of our system demonstrated by the example.

The examples presented in \texttt{Table \ref{tab:command_eg}} showcase six instances of commands generated from user requests. However, the first three examples highlight situations where our model deviates from replicating the ground truth command. A detailed analysis of these errors is provided below:

\begin{enumerate}
    \item In the first example, while the generated command achieves the desired document edit, the ground truth command exhibits more efficiency as it achieves the same outcome with fewer changes.
    \item The second example illustrates an incorrect command generated by the model, wherein it mistakes a "split" action for a "replace" action. Consequently, the edited document does not align with the intended user request.
    \item In the third example, the model considers the logo as a visual element, contrary to the ground truth, which recognizes it as a textual element within the document.
\end{enumerate}

\begin{figure*}[!htb]
\centering
\includegraphics[width=2\columnwidth]{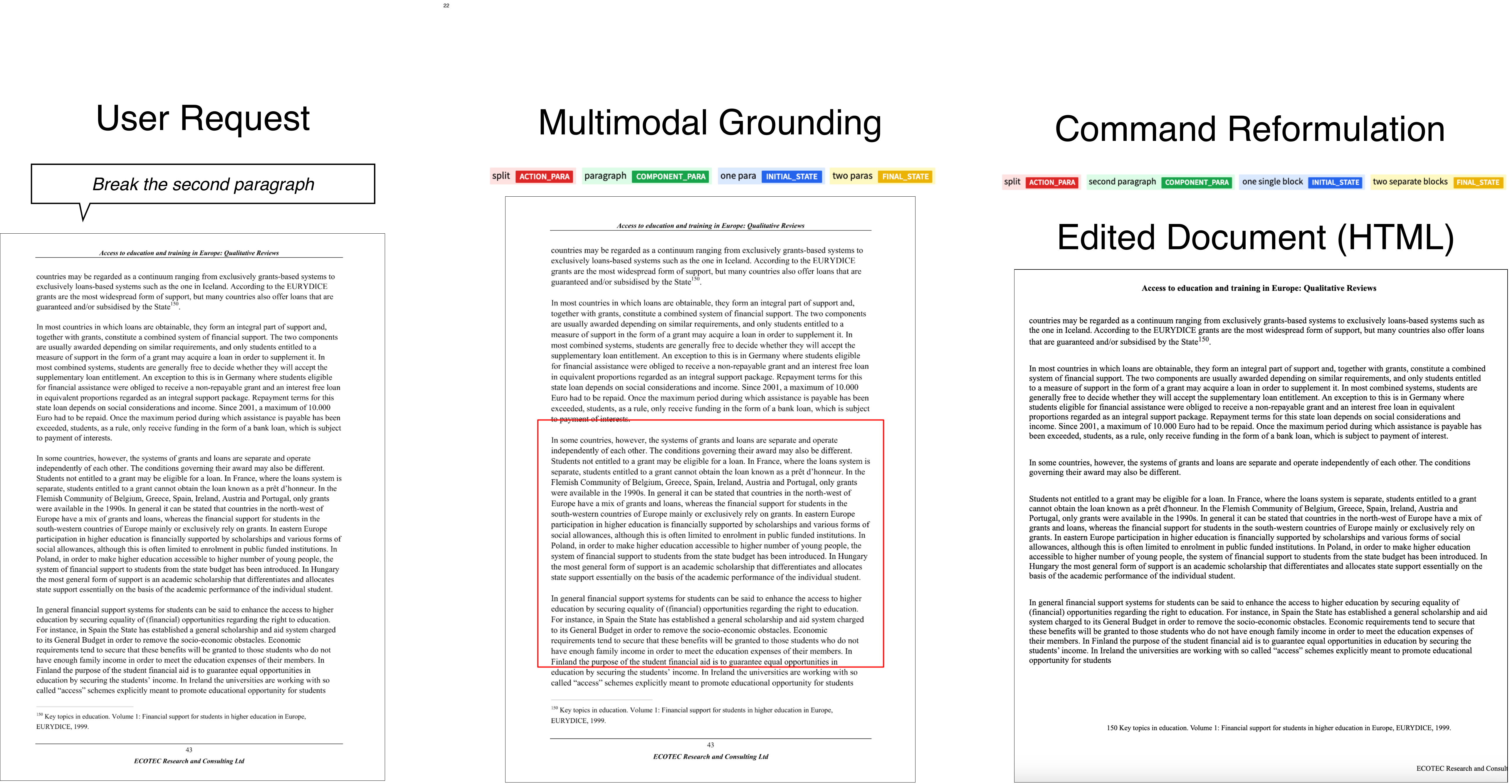}
\caption{Example of document editing request, corresponding multimodal grounding, command reformulation and edit generation.}
\label{fig:eg1}
\end{figure*}

\begin{figure*}[!htb]
\centering
\includegraphics[width=2\columnwidth]{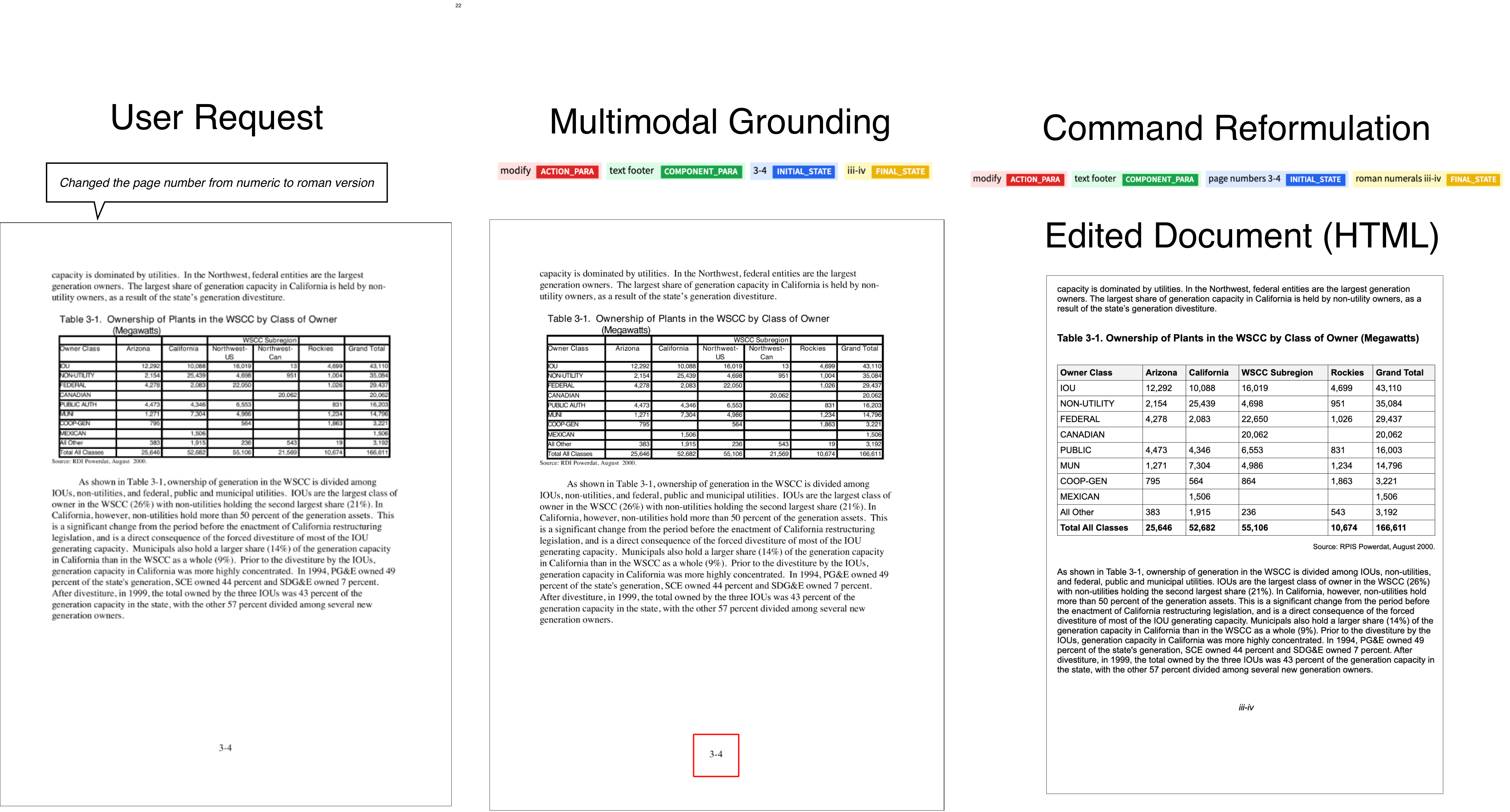}
\caption{Example of document editing request, corresponding multimodal grounding, command reformulation and edit generation.}
\label{fig:eg2}
\end{figure*}

\begin{figure*}[!htb]
\centering
\includegraphics[width=2\columnwidth]{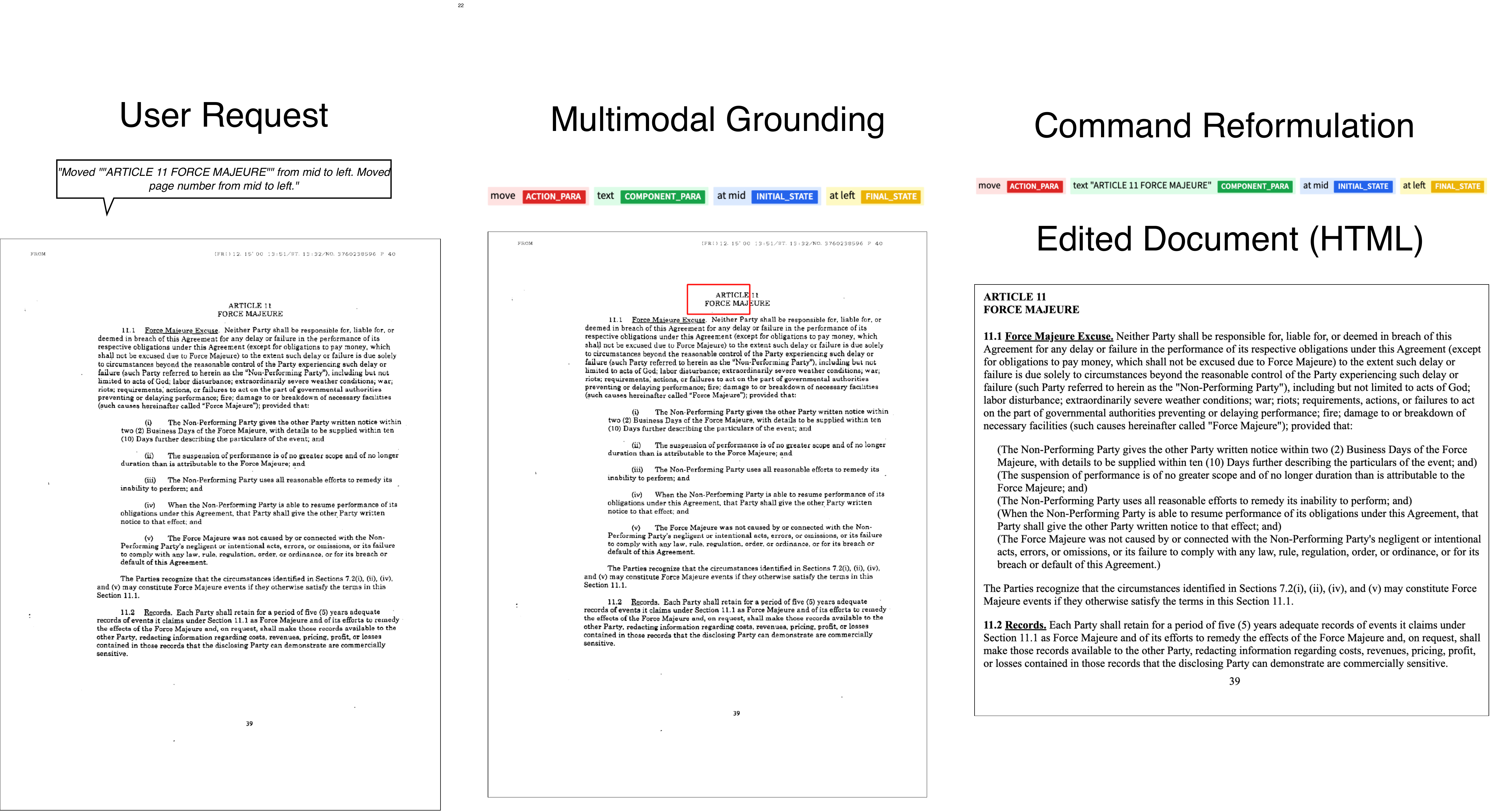}
\caption{Example of document editing request, corresponding multimodal grounding, command reformulation and edit generation.}
\label{fig:eg3}
\end{figure*}

\begin{figure*}[!htb]
\centering
\includegraphics[width=2\columnwidth]{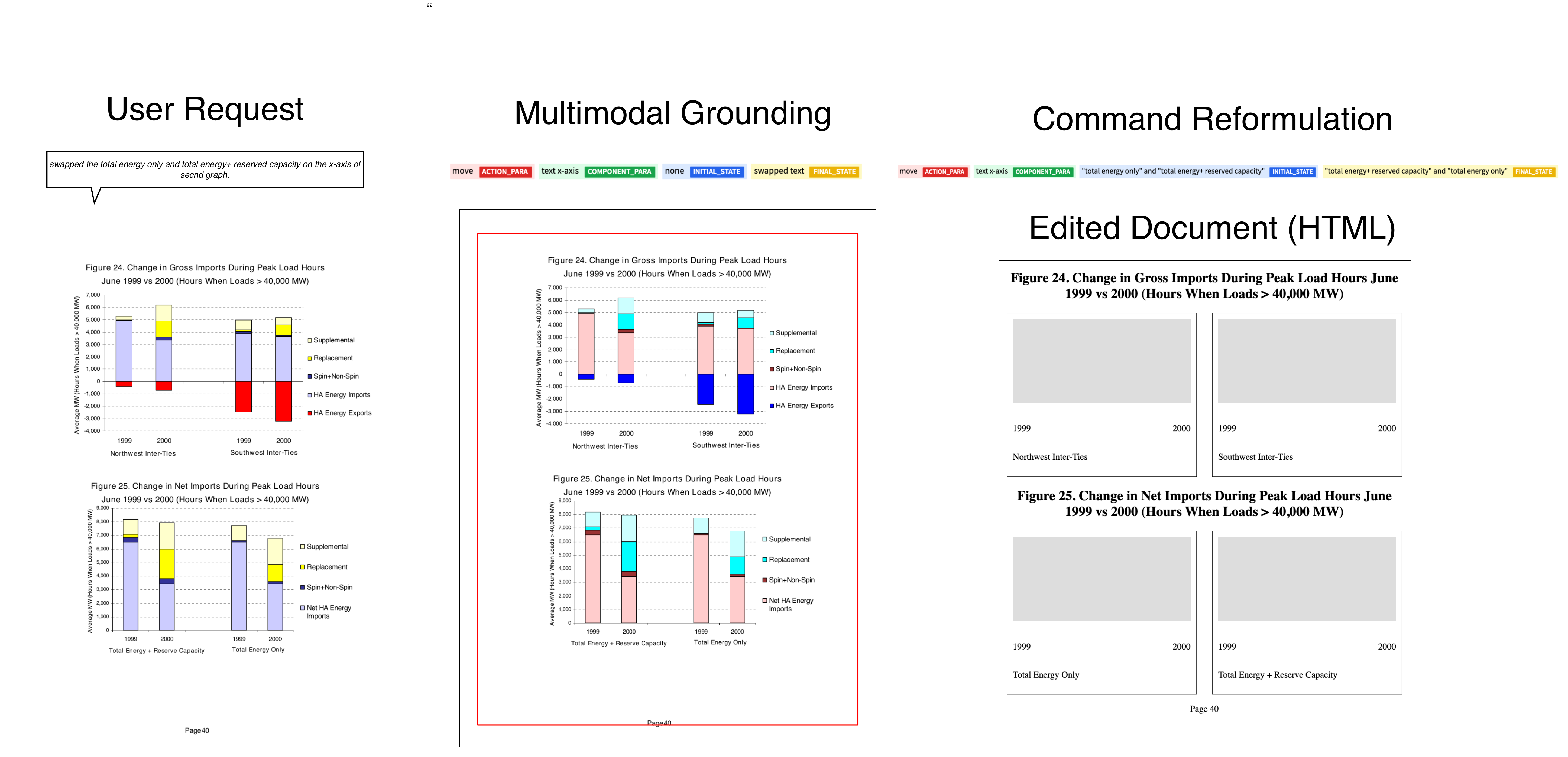}
\caption{Example of document editing request, corresponding multimodal grounding, command reformulation and edit generation.}
\label{fig:eg4}
\end{figure*}

\begin{figure*}[!htb]
\centering
\includegraphics[width=2\columnwidth]{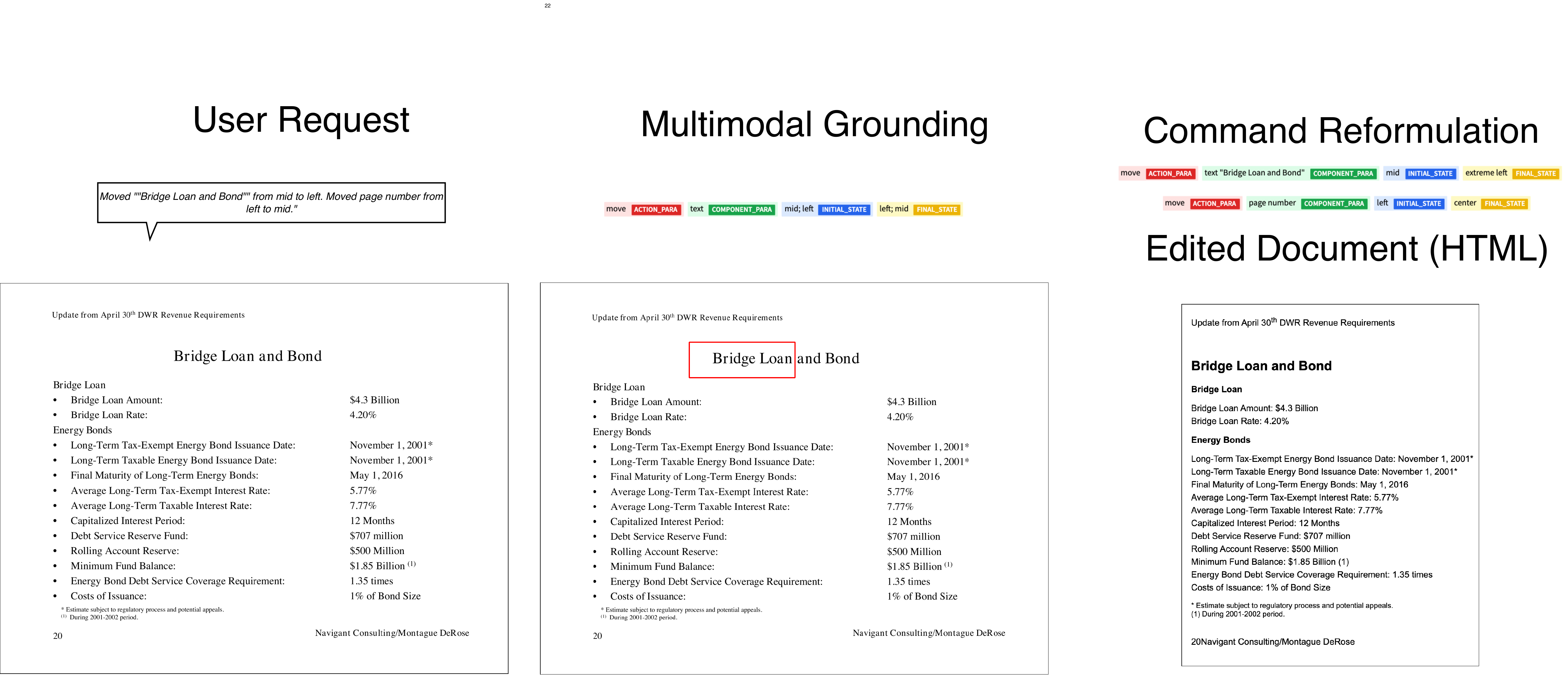}
\caption{Example of document editing request, corresponding multimodal grounding, command reformulation and edit generation.}
\label{fig:eg5}
\end{figure*}

\begin{figure*}[!htb]
\centering
\includegraphics[width=2\columnwidth]{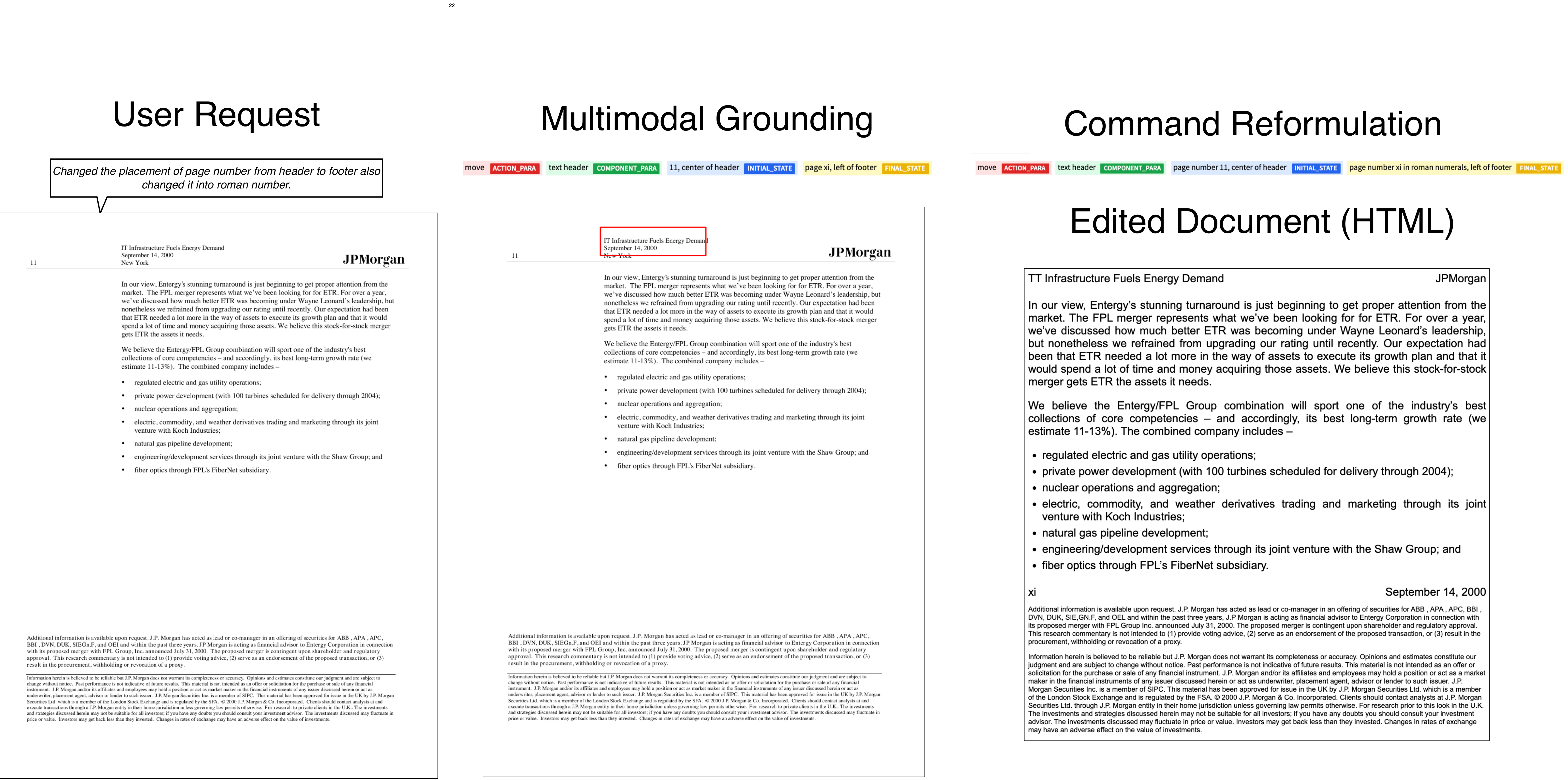}
\caption{Example of document editing request, corresponding multimodal grounding, command reformulation and edit generation.}
\label{fig:eg6}
\end{figure*}

\begin{figure*}[!htb]
\centering
\includegraphics[width=2\columnwidth]{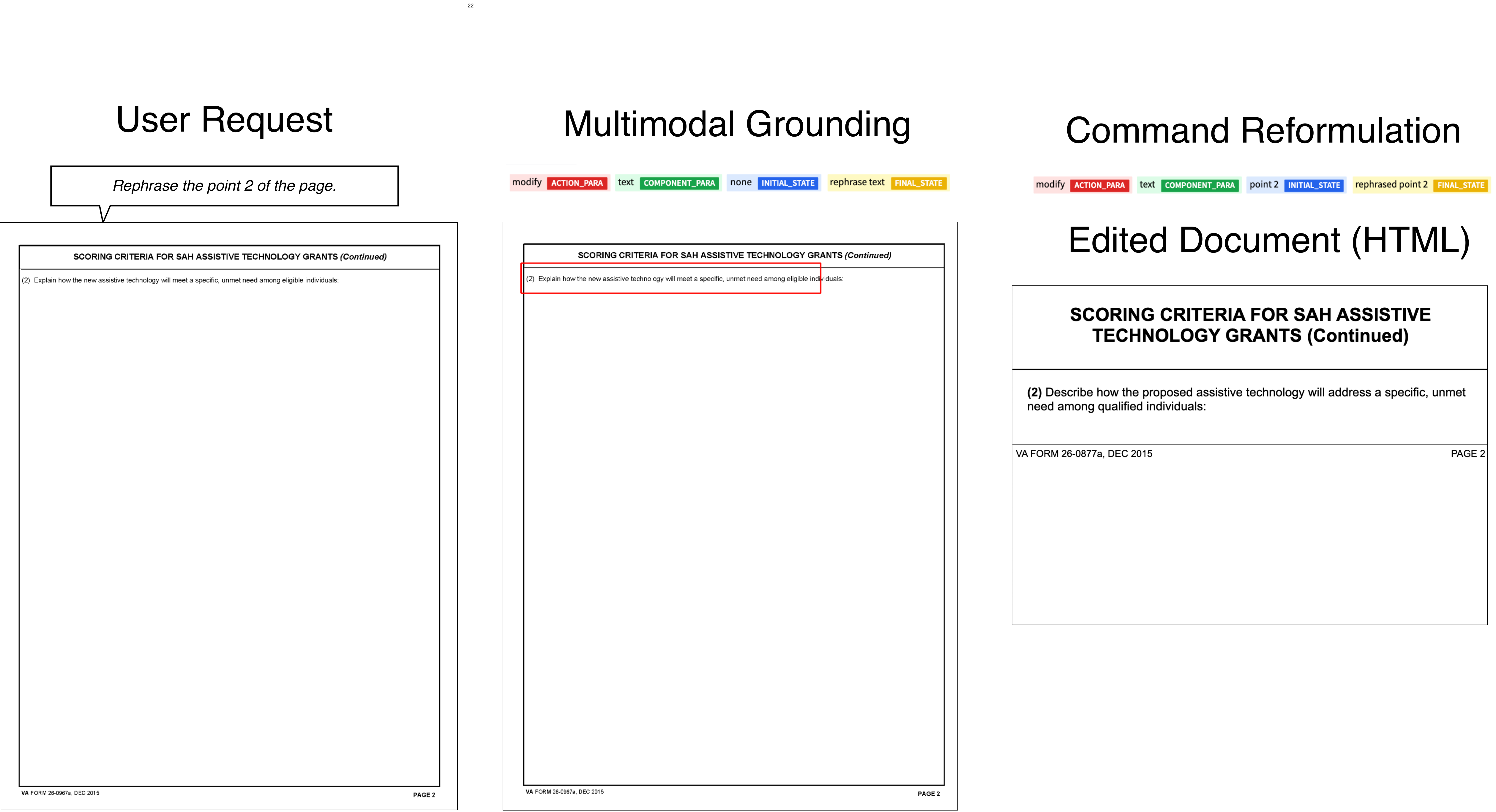}
\caption{Example of document editing request, corresponding multimodal grounding, command reformulation and edit generation.}
\label{fig:eg7}
\end{figure*}

\begin{figure*}[!htb]
\centering
\includegraphics[width=2\columnwidth]{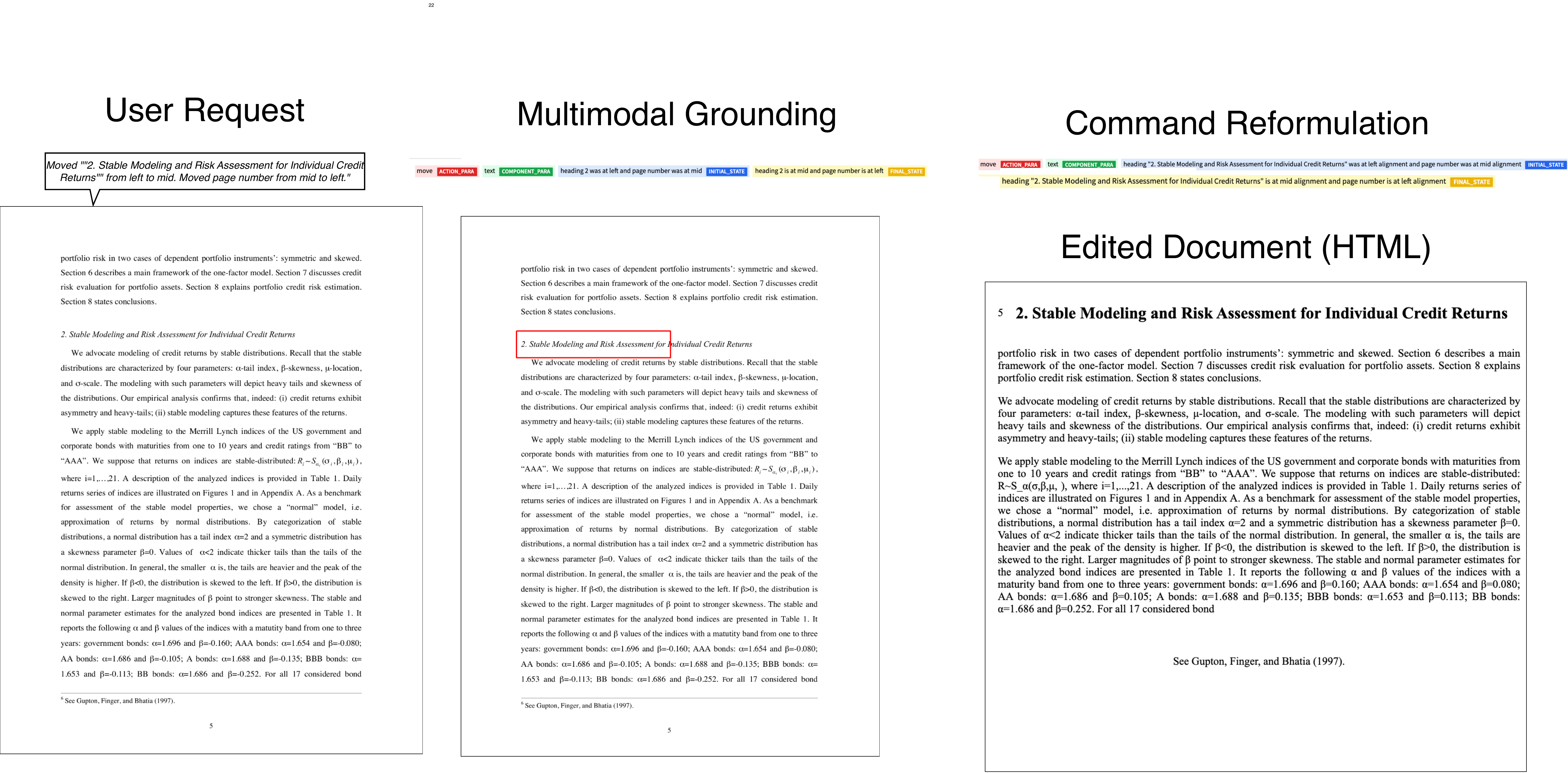}
\caption{Example of document editing request, corresponding multimodal grounding, command reformulation and edit generation.}
\label{fig:eg8}
\end{figure*}

\begin{figure*}[!htb]
\centering
\includegraphics[width=2\columnwidth]{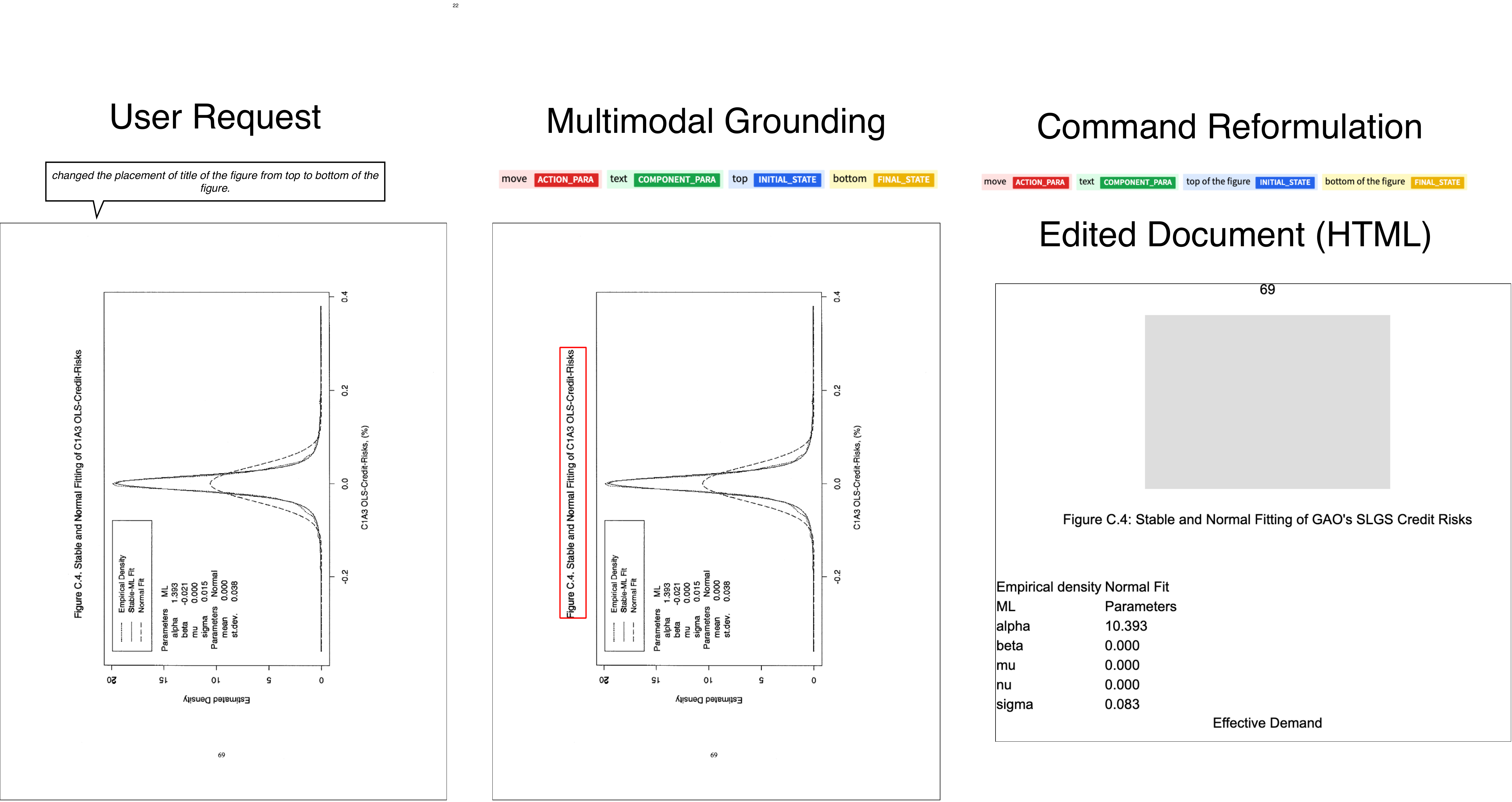}
\caption{Example of document editing request, corresponding multimodal grounding, command reformulation and edit generation.}
\label{fig:eg9}
\end{figure*}

Examples of end to end document editing are shown in Fig \ref{fig:eg1}-\ref{fig:eg9}. Each of these figures illustrates the user request and document image, followed by multimodal grounding using Doc2Command, command reformulation and finally the rendered HTML+CSS document.

% An extended demo is available at \href{youtu.be/E_dKeK82fyY}{https://youtu.be/E\_dKeK82fyY}.

% \usepackage{multirow}
% \usepackage{graphicx}
\begin{table*}[h!]
\resizebox{2\columnwidth}{!}{%
\begin{tabular}{llllll}
\hline
\textbf{User Request} &
   &
  \textbf{ACTION\_PARA} &
  \textbf{COMPONENT\_PARA} &
  \textbf{INITIAL\_STATE} &
  \textbf{FINAL\_STATE} \\ \hline
\multicolumn{1}{l|}{} &
  \multicolumn{1}{l|}{\textit{Predicted}} &
  \cellcolor[HTML]{FFCCC9}replace &
  \cellcolor[HTML]{9AFF99}text &
  \multicolumn{1}{r}{\cellcolor[HTML]{FFCCC9}December 1, 2000} &
  \cellcolor[HTML]{FFCCC9}December, 11, 2000 \\
\multicolumn{1}{l|}{\multirow{-2}{*}{Change the date "December 1, 2000" to December 11, 2020}} &
  \multicolumn{1}{l|}{\textit{Ground Truth}} &
  modify &
  text &
  1, 2000 &
  11, 2000 \\ \hline
\multicolumn{1}{l|}{} &
  \multicolumn{1}{l|}{\textit{Predicted}} &
  \cellcolor[HTML]{FFCCC9}replace &
  \cellcolor[HTML]{FFCCC9}bullet &
  \cellcolor[HTML]{FFCCC9}{\color[HTML]{222222} dotted} &
  \cellcolor[HTML]{FFCCC9}{\color[HTML]{222222} 4 bullet points} \\
\multicolumn{1}{l|}{\multirow{-2}{*}{\begin{tabular}[c]{@{}l@{}}2-3 lines of text in the paragraph "(p) Issues, obtain" are changed to four separate bullet points. Bullet a. "any department or agency \\ of the United States", b."from other agencies of the state",  c. "from any private company"  and d. "any insurance or guarantee to"\end{tabular}}} &
  \multicolumn{1}{l|}{\textit{Ground Truth}} &
  split &
  text &
  paragraph &
  split \\ \hline
\multicolumn{1}{l|}{} &
  \multicolumn{1}{l|}{\textit{Predicted}} &
  \cellcolor[HTML]{9AFF99}move &
  \cellcolor[HTML]{FFCCC9}image &
  \cellcolor[HTML]{9AFF99}left &
  \cellcolor[HTML]{9AFF99}right \\
\multicolumn{1}{l|}{\multirow{-2}{*}{Moved logo from left to right.}} &
  \multicolumn{1}{l|}{\textit{Ground Truth}} &
  move &
  text &
  left &
  right \\ \hline
\multicolumn{1}{l|}{} &
  \multicolumn{1}{l|}{\textit{Predicted}} &
  \cellcolor[HTML]{9AFF99}delete &
  \cellcolor[HTML]{9AFF99}text &
  \cellcolor[HTML]{9AFF99}in table &
  \cellcolor[HTML]{9AFF99}removed \\
\multicolumn{1}{l|}{\multirow{-2}{*}{\begin{tabular}[c]{@{}l@{}}Delete all data from table "Tabela 15 Układ pasywów bilansu jednostek, z wyłączeniem banków----"\end{tabular}}} &
  \multicolumn{1}{l|}{\textit{Ground Truth}} &
  delete &
  text &
  in table &
  removed \\ \hline
\multicolumn{1}{l|}{} &
  \multicolumn{1}{l|}{\textit{Predicted}} &
  \cellcolor[HTML]{9AFF99}add &
  \cellcolor[HTML]{9AFF99}text footer &
  \cellcolor[HTML]{9AFF99}none &
  \cellcolor[HTML]{9AFF99}Page 4 \\
\multicolumn{1}{l|}{\multirow{-2}{*}{Added page number 4 at the footer of the page.}} &
  \multicolumn{1}{l|}{\textit{Ground Truth}} &
  add &
  text footer &
  none &
  Page 4 \\ \hline
\multicolumn{1}{l|}{} &
  \multicolumn{1}{l|}{\textit{Predicted}} &
  \cellcolor[HTML]{9AFF99}merge &
  \cellcolor[HTML]{9AFF99}text &
  \cellcolor[HTML]{9AFF99}not merged &
  \cellcolor[HTML]{9AFF99}merged; heading with text \\
\multicolumn{1}{l|}{\multirow{-2}{*}{removed the space after the heading fundamental corrective measures.}} &
  \multicolumn{1}{l|}{\textit{Ground Truth}} &
  merge &
  text &
  not merged &
  merged; heading with text \\ \hline
\end{tabular}%
}
\caption{Examples of command generation in Doc2Command. Correct command parameters are highlighted in green, and incorrect command parameters are highlighted in red.}
\label{tab:command_eg}
\end{table*}

\subsection{Prompt Templates}
\begin{figure}[!htb]
\centering
\includegraphics[width=\columnwidth]{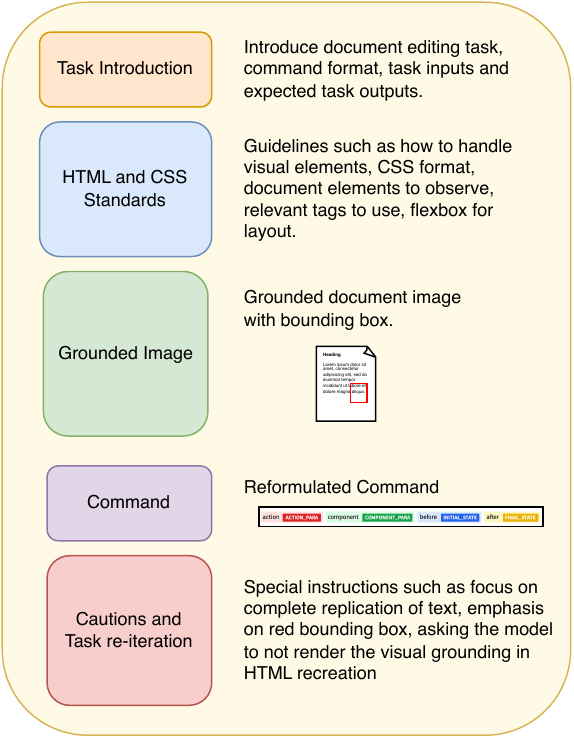}
\caption{Template of prompt used for document editing using a suitable LMM and multimodally grounded edit request.}
\label{fig:prompt1}
\end{figure}

\begin{figure}[!htb]
\centering
\includegraphics[width=\columnwidth]{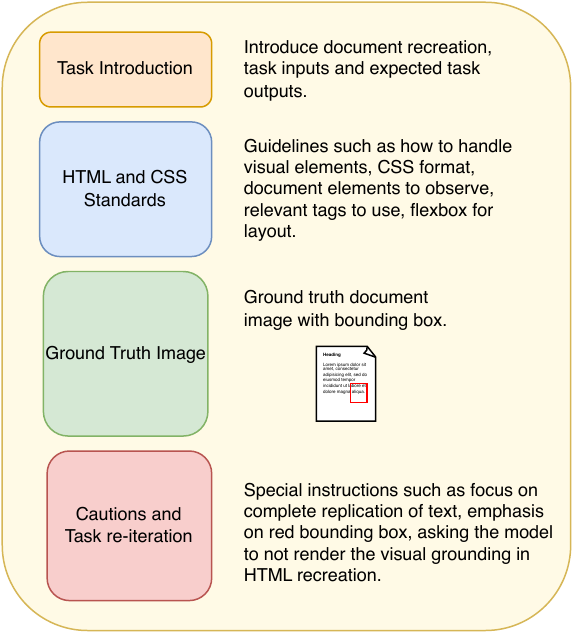}
\caption{Template of prompt used for generating ground truth document edits from post-edit, visually grounded document images.}
\label{fig:prompt2}
\end{figure}

\begin{figure}[!htb]
\centering
\includegraphics[width=\columnwidth]{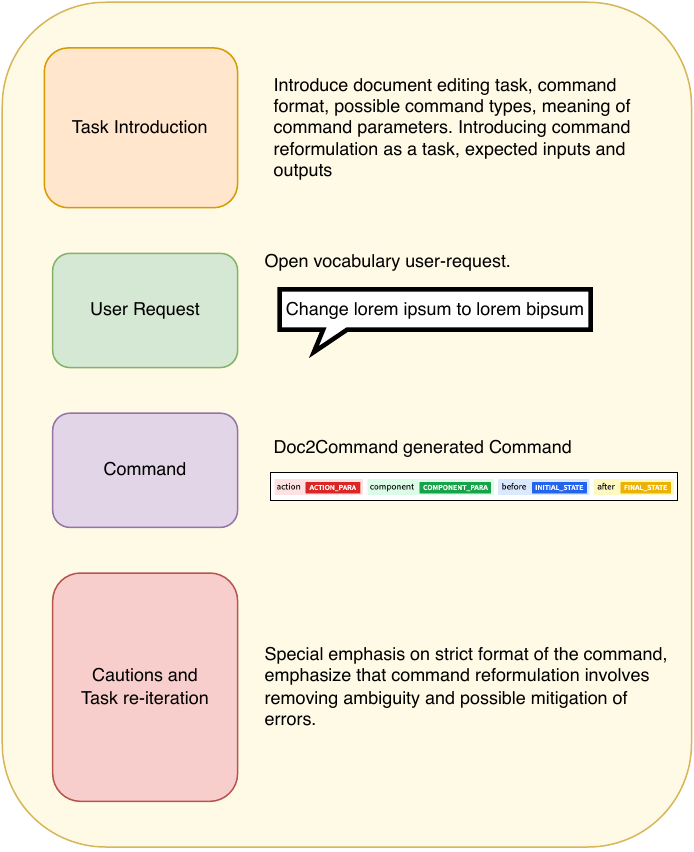}
\caption{Template of prompt used for reformulating the Doc2Command generated command using an LLM.}
\label{fig:prompt3}
\end{figure}

Fig \ref{fig:prompt1}, \ref{fig:prompt2} and \ref{fig:prompt3} represent the prompt templates used in different steps of our pipeline, with Large Language Models or Large Multimodal Models.

\subsection{Additional Evaluation Metrics}
\label{sec:eval_metrics}

We adapt these metrics from \cite{mathur2023docedit}.

\textbf{Command Grounding Metrics}
\begin{itemize}
    \item Exact Match: Percentage of generated commands that exactly match the ground truth commands.
    \item Word Overlap F1: Measures the F1 of the word overlap score between the generated and ground truth commands.
    \item ROUGE-L: Evaluates the longest common subsequence of words between the generated and ground truth commands.
    \item Action (\%): Percentage of commands with exact matches in the action parameter.
    \item Component (\%): Percentage of commands with exact matches in the component parameter.
\end{itemize}

\textbf{Visual Grounding Metrics}
\begin{itemize}
    \item Top-1 Accuracy: Measures the accuracy of visual grounding, where a match is considered when the Jaccard overlap is greater than or equal to 0.5.
\end{itemize}

\subsection{Computational Resources}
\label{sec:compute}
Table \ref{tab:comp_resources} gives an overview of computational resources used in our experiments for Doc2Command. 
\begin{table}
    \centering
    \resizebox{0.8\columnwidth}{!}{
    \begin{tabular}{c|c}
        \hline
        \textbf{Parameter} & \textbf{Value} \\
        \hline
        GPU Hours & 100 \\
        Number of Parameters & 300M \\
        GPU Specification & NVIDIA GeForce RTX 2080 Ti \\
        Number of GPUs & 1 \\
        \hline
    \end{tabular}}
    \caption{Overview of computational resources required in training and experimenting with Doc2Command.}
    \label{tab:comp_resources}
\end{table}

\subsection{Human Evaluation Instructions}
\label{sec:eval_instructions}
\begin{figure*}
    \centering
    \includegraphics[width=2\columnwidth]{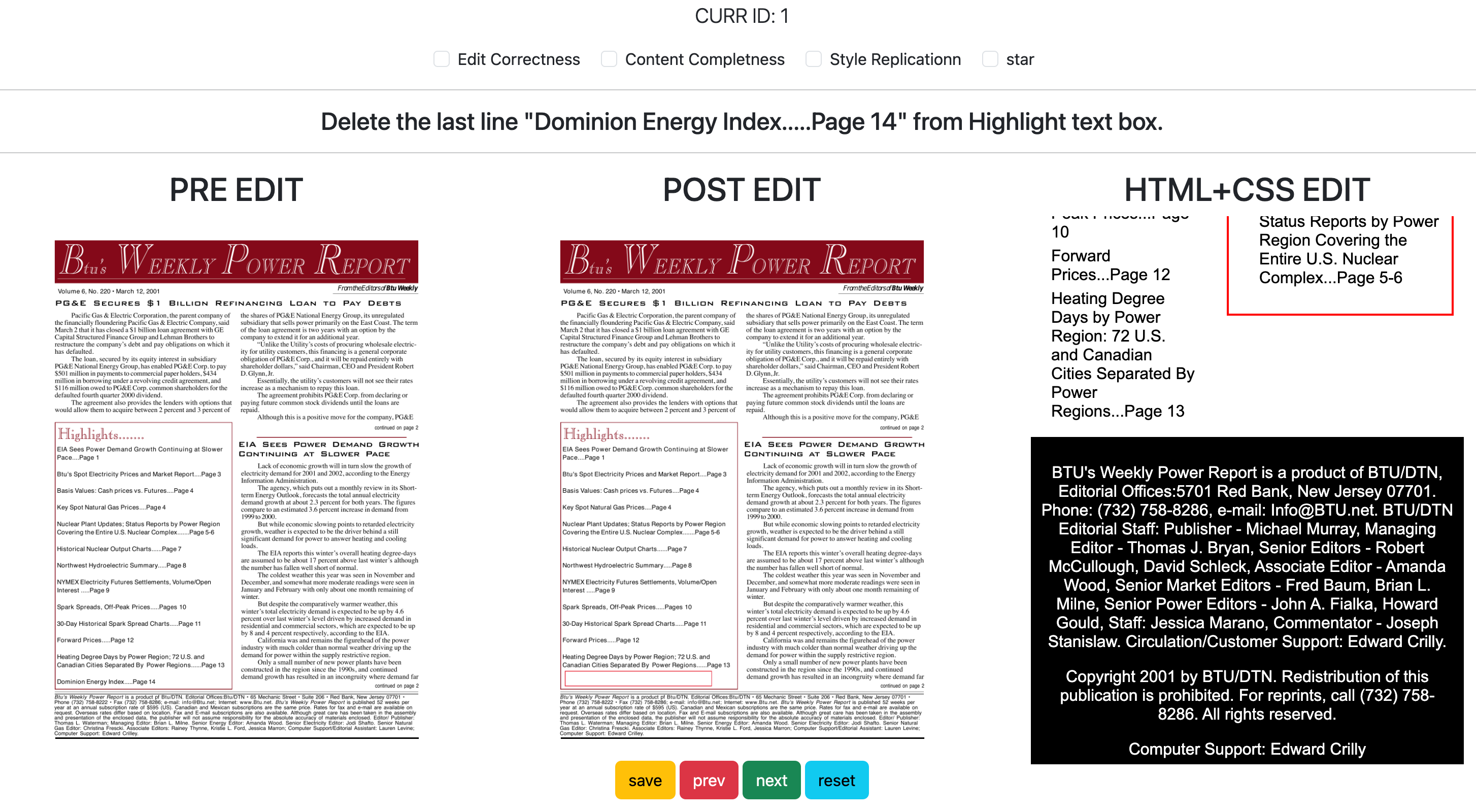}
    \caption{UI used by annotators for human evaluation.}
    \label{fig:UI}
\end{figure*}

The human evaluators are college graduates expected to have basic knowledge of working with PDF documents. They are provided with a comprehensive rubric for evaluation and a set of examples to guide to demonstrate the evaluation process. Fig \ref{fig:UI} shows the UI used by human evaluators, and table \ref{tab:human_eval} shows a concise version of the evaluation rubric annotators are expected to refer for each sample. Each annotator examines the renderings of the edited HTML document generated by \texttt{DocEdit-v2} and the ground truth pre- and post-edit document images. Evaluator are compensated well above average wages according to their geographical locations for their contributions.

\begin{table*}[ht]
\centering
\resizebox{1.3\columnwidth}{!}{
\begin{tabular}{|p{5cm}|p{10cm}|}
\hline
\textbf{Option} & \textbf{Criteria} \\ \hline
\textbf{Content Replication} & 
You should check the Content Completeness (score=1) option if \textbf{all} of the following apply:
\begin{itemize}[leftmargin=1.5em]
    \item[\cmark] Elements to be modified are included in the recreation.
    \item[\cmark] At least ~80\% of textual content has been included in the recreation.
    \item[\cmark] Visual content like figures or charts, if present in the original document are supplanted by placeholders.
\end{itemize}
Further, you should not check the Content Completeness (score=0) option if \textbf{any} of the following apply:
\begin{itemize}[leftmargin=1.5em]
    \item[\xmark] Elements to be modified are not included in the recreation.
    \item[\xmark] If the model replaces original text with fillers like\textit{ Lorem Ipsum} or hallucinates the document text by a margin of >~20\%.

\end{itemize}Refer to the example set in case of any confusion to understand different case scenarios for Content Completeness. \\ \hline
\textbf{Style Replication} & 
You should check the Style Replication (score=1) option if \textbf{most} of the following apply:
\begin{itemize}[leftmargin=1.5em]
    \item[\cmark] Layout of the elements is correct.
    \begin{itemize}
        \item[\cmark] Number of columns the page is divided into.
        \item[\cmark] Position of the text blocks is correct.
        \item[\cmark] Presence of headers/footers.
        \item[\cmark] Alignment and relative placement of elements like dates, page numbers, headings, etc.
    \end{itemize}
    \item[\cmark] Relative text size of different elements is correct. (Example: headings are larger than the text).
    \item[\cmark] Special text like bold/italics/highlight/underline is consistent with the original document.
    \item[\cmark] Relevant elements such as tables, lists or form elements have been used in HTML for document recreation.
\end{itemize} Each sample contains numerous elements, so you must verify if these rules apply to every individual element before making a decision on if a significant majority of elements are correctly styled. Please refer to the provided example set to understand the acceptable level of deviation for a document to receive a score of 1 for Style Replication.
 \\ \hline
\textbf{Edit Correctness} &
Carefully review the edit request and examine the pre-change document image. As an annotator, your task is to evaluate what the desired change should look like based on the provided instructions. Pay close attention to specific details and elements mentioned in the request. Consider the overall context and purpose of the document to ensure that your interpretation aligns with the user's intention. By thoroughly understanding the pre-change state and the requested modifications, you will be able to accurately assess the changes and ensure they are implemented correctly. This detailed evaluation is crucial for maintaining the quality and consistency of the document.
You should check the Edit Correctness (score=1) option if the following apply:
\begin{itemize}[leftmargin=1.5em]
    \item[\cmark] Changes made in the region of interest marked in the ground truth post-edit document image have been EXACTLY replicated in the HTML+CSS rendering.
    \item[\cmark] Changes made in the HTML+CSS rendering are consistent with the original user request.
\end{itemize} 
Dealing with conflicts:
\begin{itemize}[leftmargin=1.5em]
    \item[\cmark] Ambiguous user intention: change is consistent with the user request (i.e. naively fulfills the expectation) but not exactly the same as the ground truth post-edit image.
    \begin{itemize}
        \item Examples of such conflicts include: element to be modified is ambiguous, or desired change can be reasonably interpreted in multiple ways, score it as 1.
    \end{itemize}
    \item[\xmark] Incomplete modification: If the modified HTML+CSS document implements a modification that does not complete the scope of the original document request or doesn’t reasonably replicate the changes demonstrated in the ground truth post-edit document image, score it as 0.
\end{itemize}

\\ \hline
\textbf{Star} & Use the star option if a sample is extremely hard to annotate under any of the above-mentioned categories (low confidence examples) OR if the example demonstrates a unique capability of our document editing system. \\ \hline
\end{tabular}}
\caption{Concise Evaluation Criteria for Human Evaluation}
\label{tab:human_eval}
\end{table*}

\subsection{Methodology: Doc2Command Command Generation}
\label{sec:methodology_doc2command}
The input image is represented as $I \in \mathbb{R}^{H \times W \times C}$, where $H$ and $W$ are the re-scaled height and width of the image, and $C$ is the number of channels. To prepare the image as input into the transformer style encoder, the image is divided into patches, denoted by $P_{i,j} \in \mathbb{R}^{p \times P \times C}$, where $p$ is the patch size and $i, j$ index the patches. Each patch is flattened to obtain a vector of pixel values: $V_{i,j} \in \mathbb{R}^{P^2 \times C}$. The flattened patches are then fed into the image encoder ($\mathcal{E}_{I}$) to generate patch encodings $Z_I = \{Z_{i,j} \forall i,j\}, Z_I \in \mathbb{R}^{N \times d1}$ such that $Z_{i,j} = \mathcal{E}_{I}(V_{i,j})$, where $N$ is the number of patches and $d1$ is the encoder dimension. The patch embeddings generated by the encoder serve as input to the text decoder, which auto-regressively generates a sequence of $r$ tokens, $CT$ representing the command text as $CT = \mathcal{D}_{T}(Z)$, where $CT = \{s_1, s_2 \dots s_r\}$. The taxonomy of actions includes Add, Delete, Copy, Move, Replace, Split, Merge, and Modify.

\subsection{Methodology: Doc2Command Multimodal Grounding}
\label{sec:methodology_grounding}.
A point-wise linear layer is applied to the patch encoding $Z \in \mathbb{R}^{N \times D}$ to produce patch-level class logits $Z_{\text{lin}} \in \mathbb{R}^{N \times K}$. The sequence is then reshaped into a 2D feature map $S_{\text{lin}} \in \mathbb{R}^{H/P \times W/P \times K}$ and bilinearly upsampled to the original image size $S \in \mathbb{R}^{H \times W \times K}$. A softmax is applied to the class dimension to obtain the final segmentation map. A set of learnable class embeddings $C \in \mathbb{R}^{K \times d_2}$ is introduced, where $K$ is the number of classes ($K=3$ for our model), and $d_2$ is the mask-transformer dimension. Each class embedding is initialized randomly and assigned to a single semantic class. It is used to generate the class mask. The mask-transformer processes the class embeddings jointly with patch encodings $Z_I \in \mathbb{R}^{N \times D}$ such that $C, Z_M = \mathcal{D}_I(C_0, Z_I)$. The mask transformer generates $K$ masks by computing the scalar product between L2-normalized patch embeddings $Z_M \in \mathbb{R}^{N \times d2}$ and class embeddings $C \in \mathbb{R}^{K \times d2}$ output by the decoder as $\mathcal{M}_I = Z_M \cdot C^T$. The set of class masks is reshaped into a 2D mask $S_{I} \in \mathbb{R}^{H/P \times W/P \times K}$ and bilinearly upsampled to the image size to obtain a feature map $S \in \mathbb{R}^{H \times W \times K}$. A softmax is then applied to the class dimension, followed by layer normalization to obtain pixel-wise class scores, forming the final segmentation map. The mask sequences are softly exclusive to each other, i.e., $\sum_{k=1}^{K} S_{i,j,k} = 1$ for all $(i, j) \in H \times W$.
The Region of Interest (RoI) is represented by the bounding box \( [x, y, h, w] \). 

% \subsection{More Results}
% \label{sec:more_results}
% Fig. \ref{fig:corr} compares multimodal grounding metrics and human evaluation for document editing. Intuitively, the document edit is sub-optimal when the generated command is incorrect. However, since we have a command optimization module, Fig. \ref{fig:corr} (a) shows a wide spread of high-scored document editing, even with initially incorrect Doc2Command commands. Fig. \ref{fig:corr} (b) plots visual grounding IoU with the document edit score and demonstrates the crucial role of visual grounding in document editing.

% \begin{figure}[h!]
%     \centering
%     \begin{subfigure}{0.23\textwidth}
%         \centering
%         \includegraphics[width=\textwidth]{em_scatter.pdf}
%         \caption{Comparison between the command text EM value and human evaluation scores.}
%         \label{fig:em_scatter}
%     \end{subfigure}
%     \hfill
%     \begin{subfigure}{0.23\textwidth}
%         \centering
%         \includegraphics[width=\textwidth]{iou_scatter.pdf}
%         \caption{Comparison between the visual grounding IoU and human evaluation scores.}
%         \label{fig:eg2}
%     \end{subfigure}
%     \caption{Plotting the correlation between document editing performance and multimodal grounding
%     .}
%     \label{fig:corr}
% \end{figure}
\end{document}